%% file: iclr2027_conference.tex
\documentclass{article}
\usepackage{lineno}
\usepackage{iclr2027_conference,times}

\input{math_commands.tex}

\usepackage{graphicx}
\usepackage{booktabs}
\usepackage{multirow}
\usepackage[table]{xcolor}
\usepackage{amssymb}
\usepackage{bbm}
\usepackage{hyperref}
\definecolor{linkcyan}{RGB}{0,173,239}
\definecolor{linkred}{RGB}{204,0,0}
\hypersetup{
  colorlinks = true,
  linkcolor  = linkred,
  citecolor  = linkcyan,
  urlcolor   = linkcyan,
}
\usepackage{url}
\usepackage{wrapfig}
\usepackage{capt-of}
\definecolor{rowblue}{RGB}{210, 235, 253}

\newcommand{\blanksymbolfootnote}[1]{%
  \renewcommand{\thefootnote}{}% Remove footnote number
  \footnote{#1}%
  \setcounter{footnote}{0} % Reset footnote counter
  \renewcommand{\thefootnote}{\arabic{footnote}}% Restore footnote number
}

\title{ScopeIF: Improving Scope-Aware Precise Instruction-Following in Large Language Models via Graded Reward Modeling}

\author{%
  \parbox[t]{\dimexpr\textwidth-2\tabcolsep\relax}{%
    \normalfont
    \textbf{Bosi Wen$^{1,\dagger}$}, 
    \textbf{Yilin Niu$^{2}$}, 
    \textbf{Xiaoying Ning$^{2}$}, 
    \textbf{Ying Zhang$^{2}$}, 
    \textbf{Hongning Wang$^{1}$}, 
    \textbf{Minlie Huang$^{1,\ddagger}$}
    }\\
    $^{1}$The Conversational Artificial Intelligence (CoAI) Group, Tsinghua University \\
    $^{2}$Zhipu AI \\
    \texttt{wbs23@mails.tsinghua.edu.cn, aihuang@tsinghua.edu.cn}
  }%
\iclrfinalcopy

\begin{document}
\maketitle
\lhead{Under review}

\blanksymbolfootnote{$^\dagger$Work done when this author interned at Zhipu AI.}
\blanksymbolfootnote{$^\ddagger$Corresponding author}

\begin{abstract}
\input{sections/1_abstract}
\end{abstract}

\section{Introduction}
\input{sections/2_introduction}

\section{Related Work}
\input{sections/3_related_work}

\section{Preliminaries}
\input{sections/4_preliminaries}

\section{Data Construction}
\input{sections/5_data_construction}

\section{Methodology}
\input{sections/6_methodology}

\section{Experiments}
\input{sections/7_experiments}

\section{Conclusion}
\input{sections/8_conclusion}

% \section*{Limitations}
% \input{sections/limitation}

\section*{AI use statement}
In this work, we used generative AI tools to generate synthetic datasets and to clean and reformat these data.
We have not used them for any other required-disclosure task. 
Additionally, we used generative AI tools to polish our figures and manuscript. 
We have reviewed all AI-assisted work, verifying the generated data through our quality-control pipeline and checking all AI-assisted figures and text ourselves. 
We take responsibility for the final content of this work, including text, claims, or artifacts produced with the aid of generative AI.

\section*{Reproducibility Statement}
To ensure the reproducibility of our experiments, we have released the full implementation of our ScopeIF framework (\url{https://github.com/thu-coai/ScopeIF}), which includes full
data and code for data collection and RL training. 
The prompts applied throughout this work are provided
in Appendix \ref{app:prompt_template}. 
Crucial hyperparameters for training our models and the baselines are documented
in Appendix \ref{app:training_details}.

\bibliography{custom}
\bibliographystyle{iclr2027_conference}

\clearpage
\appendix
\input{sections/9_appendix}

\end{document}

%% file: math_commands.tex
\usepackage{amsmath,amsfonts,bm}

\def\eqref#1{equation~\ref{#1}}
\def\1{\bm{1}}

\DeclareMathAlphabet{\mathsfit}{\encodingdefault}{\sfdefault}{m}{sl}
\SetMathAlphabet{\mathsfit}{bold}{\encodingdefault}{\sfdefault}{bx}{n}

%% file: sections/1_abstract.tex
Precise instruction-following is a fundamental ability of large language models (LLMs), requiring their outputs to strictly satisfy objective constraints in input instructions. 
In complex application scenarios, these constraints often possess diverse scopes that govern specific response segments rather than the entire output. 
However, existing optimization methods often neglect constraint scope during data construction and rely on binary per-constraint rewards, yielding limited data diversity and sparse supervision for complex constraints. 
To this end, we propose ScopeIF, a novel training framework for scope-aware precise instruction-following. 
We first introduce a unified schema that factorizes objective constraints into three decoupled dimensions: \textit{Scope}, \textit{Target}, and \textit{Range}. 
Grounded in this schema, we construct \textsc{ScopeInstruct}, a large-scale instruction dataset with diverse scope-aware constraints, and combine tool-grounded verification with graded reward modeling to quantify the violation degree of each constraint, providing dense supervision for policy optimization. 
Extensive experiments demonstrate that ScopeIF consistently outperforms existing methods, particularly on complex scope-aware constraints, while preserving general capabilities.
Notably, it enables optimized Qwen3-4B and 8B models to rival or surpass strong frontier models such as Gemini-2.5-Pro and DeepSeek-V3.2, establishing an effective paradigm for advancing scope-aware instruction-following.
Our code and data are available at \url{https://github.com/thu-coai/ScopeIF}.

%% file: sections/2_introduction.tex
Large language models (LLMs) have demonstrated remarkable capabilities across various NLP tasks \citep{zhao2023survey}. 
Among these capabilities, precise instruction following is a foundational prerequisite for practical applications, requiring LLMs to accurately satisfy objective output constraints that specify length, format, and content \citep{pyatkin2026generalizing}. 
Strict adherence to these constraints ensures model reliability \citep{huang2024survey} and is particularly vital in agentic and tool-assisted workflows that demand rigorous compliance with predefined schemas. 
In such scenarios, even minor deviations can cause parsing failures and downstream system breakdowns \citep{ye2026muldimif}.

\begin{wrapfigure}{r}{0.5\linewidth}
% \begin{figure}[t]
% \scriptsize
    \vspace{-4mm}
    \centering
    \includegraphics[width=1.0\linewidth]{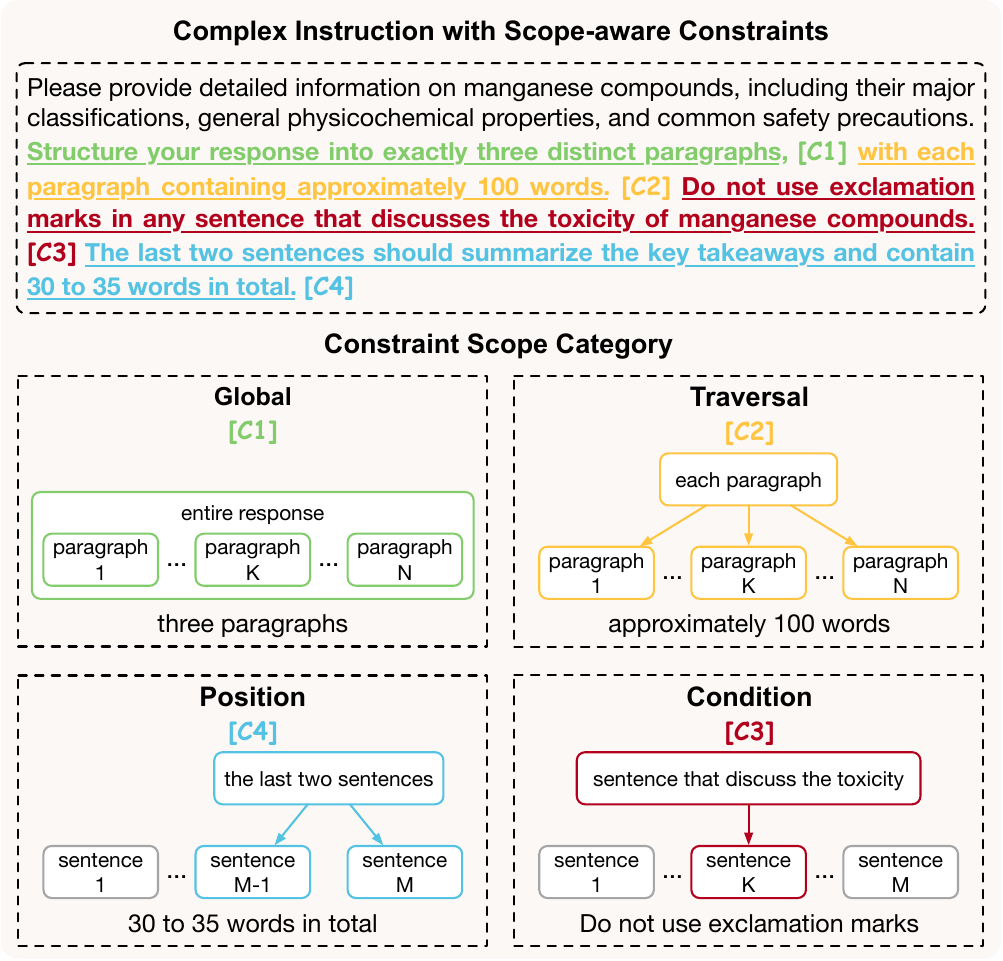}
    % \vspace{-6mm}
    \caption{An example of instruction with scope-aware constraints that govern specific response segments rather than the entire response. }
    \label{fig:example}
    % \vspace{-2mm}
% \end{figure}
\end{wrapfigure}

While current LLMs demonstrate proficiency in handling simple constraints applied to the entire response \citep{yang2025qwen3, guo2025deepseek}, they still struggle with scope-aware constraints, which govern specific segments of a response \citep{pyatkin2026generalizing, mao2026ifhierbench}, as shown in Figure \ref{fig:example}. 
These constraints require fine-grained control and meticulous planning, posing a greater challenge to LLMs. 
Given the prevalence of such constraints in real-world complex instructions \citep{mao2026ifhierbench}, there is a critical need for effective approaches to enhance this capability.

Recent advances in reinforcement learning (RL) provide a promising strategy to improve instruction-following \citep{peng-etal-2025-verif, liu2025recast, qin2026incentivizing}. 
They typically synthesize complex instructions with multiple constraints and optimize models with verifiable rewards.
Despite these efforts, they still face two critical limitations when handling scope-aware constraints.
First, for \textbf{data construction}, current methods primarily synthesize constraints applied to the entire response, neglecting the modeling of constraint scope \citep{he-etal-2024-complex, zhang-etal-2025-iopo, cheng2025spar, peng-etal-2025-verif}. 
Although some benchmarks attempt to include scope-aware constraints \citep{pyatkin2026generalizing, mao2026ifhierbench}, they remain restricted to a narrow set of hand-crafted, code-verifiable constraints, severely hindering diversity. 
Second, for \textbf{reward modeling}, existing methods typically conduct binary verification on individual constraints and aggregate their scores to derive the overall reward \citep{liu2025recast, zhang2025replay, qin2026incentivizing}.
However, as many scope-aware constraints govern multiple response segments (e.g., every sentence) or entail complex inter-segment dependencies (e.g., progressively increasing the word count across paragraphs) \citep{pyatkin2026generalizing}, LLMs often struggle to satisfy them perfectly, even after many attempts. 
Consequently, the binary reward tends to become highly sparse and ineffective for guiding optimization.

To bridge this gap, we propose ScopeIF, a novel training framework for scope-aware precise instruction following. 
Our core innovation is a unified schema that formalizes objective constraints into three decoupled dimensions: \textit{Scope} (governed segment), \textit{Target} (constrained object), and \textit{Range} (permissible bounds), which enables the modeling of constraint scopes and the computation of constraint violation degrees, directly addressing the aforementioned bottlenecks in data diversity and reward sparsity.
Grounded in this schema, ScopeIF is driven by three synergistic components. 
First, \textbf{factorized constraint synthesis}. 
To collect diverse scope-aware instructions, we flexibly combine the three dimensions to generate atomic constraints. 
LLM-based crossover is then applied to merge them into complex instructions with multiple constraints.
Second, \textbf{tool-grounded constraint verification}.
To precisely count elements and isolate relevant response segments for complex scope-aware constraint verification,
we empower judge models to dynamically design and execute tools based on response content, and then leverage the execution outcomes to derive fine-grained statistics of each \textit{Target}. 
This method blends the flexibility of LLM judges with the precision of rules.
Finally, \textbf{graded reward modeling}.
To obtain denser rewards for effective model optimization, we quantify the violation degree of each constraint by comparing the statistics of each \textit{Target} against the defined \textit{Range} to derive a per-constraint continuous graded reward. 
This reward is then hierarchically aggregated with the binary reward to guide RL training.
Our main contributions are summarized as follows:
\begin{itemize}
    \item We introduce a unified schema that factorizes objective constraints into three decoupled dimensions, enabling both diverse scope-aware constraint synthesis and constraint violation degree computation.
    Building on this, we construct \textsc{ScopeInstruct}, a large-scale dataset containing 17,968 complex instructions with diverse scope-aware constraints. 

    \item We develop ScopeIF, a novel RL training framework for scope-aware precise instruction-following that combines tool-grounded constraint verification with a graded reward modeling mechanism to provide reliable and dense supervision for effective RL optimization.

    \item We validate the effectiveness of ScopeIF across multiple policy models and benchmarks, yielding consistently better performance compared to existing frameworks and reward modeling paradigms, particularly on complex scope-aware constraints.
\end{itemize}

%% file: sections/3_related_work.tex
\paragraph{Precise Instruction-Following.} 
As LLMs are increasingly deployed in complex applications, the ability to precisely follow instructions with multiple objective constraints has become essential to their practical utility \citep{liu2023trustworthy, lou2024large}, driving extensive research to evaluate and enhance this capability from various perspectives, such as output length \citep{jie2024prompt, zhang2026lifebench, zhang2026larft} and format \citep{xia2024fofo, yao2025reff, wang2025verifiable}.
These works typically construct verifiable constraints and use rule-based feedback for automatic evaluation or model training.
Building on this, IFEval \citep{zhou2023instruction} extends precise instruction-following to 25 constraint categories, improving evaluation comprehensiveness. 
Furthermore, IFBench \citep{pyatkin2026generalizing} and IFHierBench \citep{mao2026ifhierbench} introduce hierarchical structures and scope modeling within constraints, posing greater challenges to LLMs. 
Nevertheless, these works confine precise instruction-following to a narrow set of hand-crafted, code-verifiable constraints. 
In contrast, our factorized synthesis method can automatically generate diverse scope-aware constraints at scale.

\paragraph{Instruction-Following Optimization.} 
Recent algorithmic innovations have significantly advanced the instruction-following capabilities of LLMs, which evolve from supervised fine-tuning \citep{sun2024conifer, qi2025constraint} to preference optimization \citep{zhang-etal-2025-iopo, cheng2025spar} and RL \citep{liu2025recast, zhang2025replay, wen2026if}. 
These approaches typically synthesize complex instructions with multiple constraints, conduct binary verification on each constraint, and aggregate their scores to guide model optimization.
Specifically, RECAST \citep{liu2025recast} and VerIF \citep{peng-etal-2025-verif} categorize constraints into soft and hard types, leveraging LLM-based and rule-based evaluation, respectively, to derive rewards for RL training. 
However, these methods neglect constraint scope modeling during data construction and rely on sparse binary reward for each constraint that cannot distinguish different degrees of constraint violation. 
ScopeIF addresses these limitations with a unified constraint schema that enables scope-aware data synthesis and a graded reward mechanism to capture partial constraint compliance, yielding denser supervision signals.

%% file: sections/4_preliminaries.tex
In this section, we define the precise instruction-following task and its standard evaluation metrics.

\paragraph{Task Definition.}
Our goal is to enhance the precise instruction-following capabilities of LLMs. 
Formally, given an instruction $x$ comprising a set of constraints $C = \{c_1, c_2, \dots, c_n\}$, 
an LLM is considered to follow the instruction if its response $y$ satisfies all constraints in $C$ \citep{zhou2023instruction}.
Precise instruction-following focuses on objective, deterministic constraints requiring no subjective semantic evaluation \citep{pyatkin2026generalizing}, which constitute the vast majority of output constraints in real-world developer prompts \citep{mao2026ifhierbench}.
Furthermore, beyond applying to the entire response, scope-aware constraints govern specific segments of the response (e.g., particular paragraphs or sentences), which is prevalent in real-world complex instructions \citep{mao2026ifhierbench}.

\paragraph{Evaluation Metrics.}
For each constraint, let the binary function $I(x, y, c_i) \in \{0, 1\}$ to denote whether a response $y$ satisfies the constraint $c_i$. 
The overall instruction-following quality of a response is typically evaluated using the following metrics \citep{zhang2025replay, peng-etal-2025-verif}:
\textbf{(1) Instruction-Level Accuracy (ILA)} reflects strict adherence to the entire instruction. 
A response $y$ is considered correct if and only if it satisfies every constraint of the instruction $x$, which is computed as: $
    \text{ILA}(x, y, C) = \prod_{c_i \in C} I(x, y, c_i)
$.
\textbf{(2) Constraint-Level Accuracy (CLA)} measures the ability to follow individual atomic constraints, calculated as the percentage of satisfied constraints:
$
    \text{CLA}(x, y, C) = \frac{1}{|C|} \sum_{c_i \in C} I(x, y, c_i)
$.
While ILA and CLA are both common reward modeling approaches for RL training, they rely on binary signals for individual constraints without measuring their violation degrees, leading to a sparse reward problem when facing complex constraints.

%% file: sections/5_data_construction.tex
\begin{figure*}[!t]
\scriptsize
    \centering
    \includegraphics[width=1.0\textwidth]{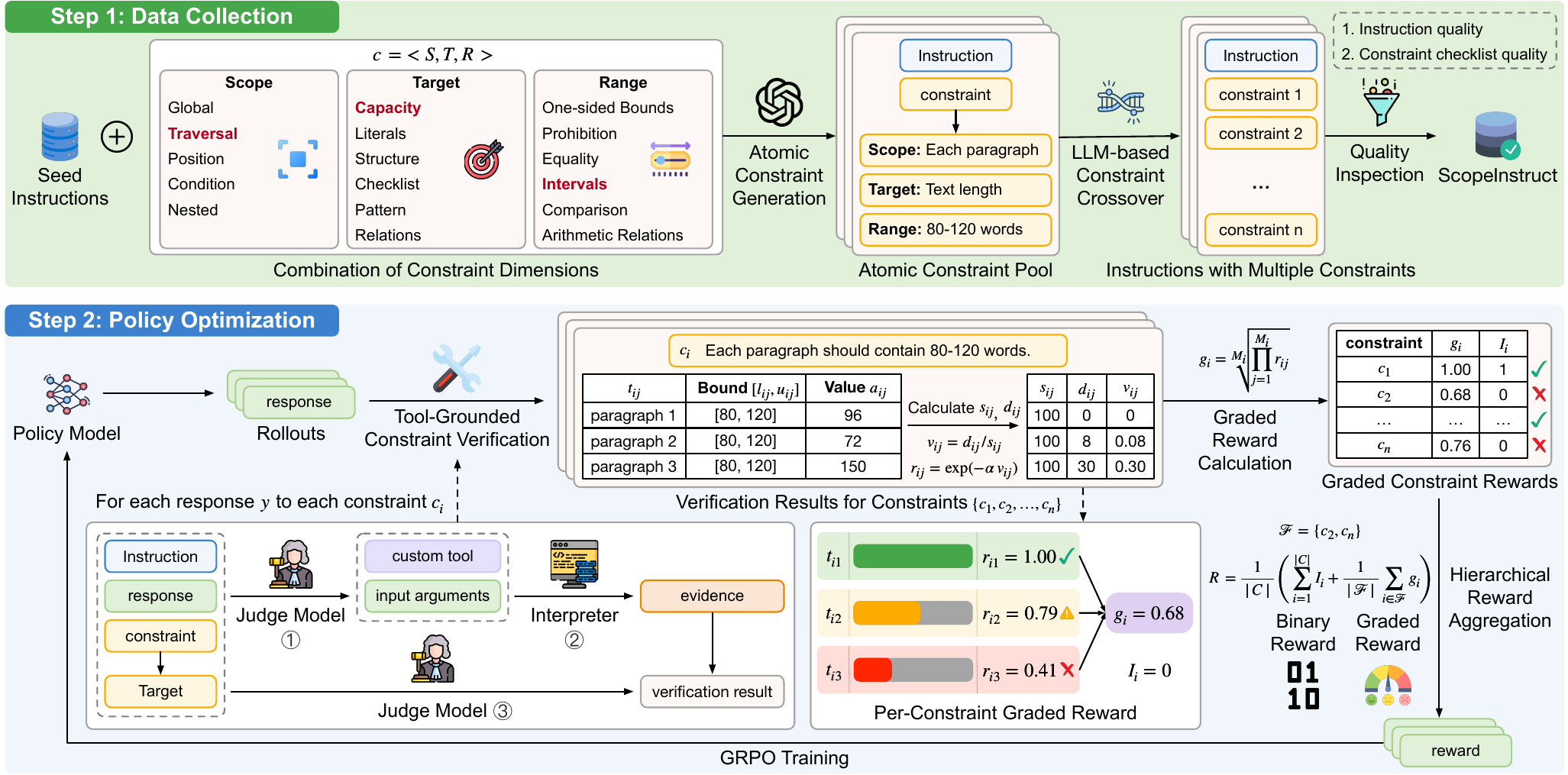}
    % \vspace{-2mm}
    \caption{Overall framework of ScopeIF. \textit{Top:} Complex instruction collection via factorized constraint synthesis. \textit{Bottom:} Policy model optimization via graded reward modeling.}
    % \vspace{-2mm}
    \label{fig:overview}
\end{figure*}

In this section, we introduce a factorized constraint synthesis method and construct \textsc{ScopeInstruct}, a large-scale dataset of complex instructions with diverse scope-aware constraints. 
We first formalize objective constraints through a unified schema comprising three dimensions: \textit{Scope}, \textit{Target}, and \textit{Range}.
Building on this, we generate diverse atomic constraints via dimension recombination and employ LLM-based crossover to synthesize instructions with multiple constraints.

\subsection{Unified Constraint Schema}
\label{sec:constraint_schema}
Existing constraint synthesis methods treat constraints as indivisible templates, confining them to a manually designed inventory and precluding a unified metric for quantifying violation degrees.
We observe that objective constraints inherently share a unified schema: they dictate the permissible bounds of specific measurable objects within designated response segments. 
Driven by this observation, we factorize each constraint $c_i$ into three decoupled dimensions: \textit{Scope}, \textit{Target}, and \textit{Range}. 
This decomposition enables diverse constraint synthesis via dimension recombination, while providing a unified basis for quantifying violation degrees.
Formally, we define this schema as follows:
\begin{equation}
\begin{aligned}
&& c_i &= \left\langle \mathcal{S}_i, \mathcal{T}_i, \mathcal{R}_i \right\rangle, \\
\mathcal{S}_i(y) &= \{\sigma_{ij}\}_{j=1}^{M_i}, \quad & \mathcal{T}_i(y) &= \{t_{ij}\}_{j=1}^{M_i}, \quad & \mathcal{R}_i(y) &= \{[l_{ij}, u_{ij}]\}_{j=1}^{M_i}.
\end{aligned}
\label{eq:factorized_constraint}
\end{equation}
Here, $\mathcal{S}_i$ denotes \textit{Scope} selector, which selects $M_i$ governed segments $\{\sigma_{ij}\}$ from response $y$ ($M_i > 1$ accommodates multi-segment scopes, such as every sentence or paragraph). 
\textit{Target} $\mathcal{T}_i$ yields the measurable object $t_{ij}$ for each segment, and \textit{Range} $\mathcal{R}_i$ assigns its permissible bounds $[l_{ij}, u_{ij}]$. 
By systematically analyzing instructions from real-world applications and existing benchmarks, we establish a comprehensive taxonomy for these three dimensions. 
More details are in Appendix \ref{app:schema}.

\paragraph{Scope} selector $\mathcal{S}_i$ isolates governed segments by applying a selection operation to response units, spanning the entire response, structural blocks and elements, paragraphs, sentences, words, and characters.
The selector falls into four categories: \textbf{(1) Global} covers the entire response, \textbf{(2) Traversal} selects every unit of a given kind (e.g., every paragraph), \textbf{(3) Position} targets units at specific locations (e.g., the beginning), and \textbf{(4) Condition} filters units through predicates (e.g., sentences containing a specific keyword).
These operations can be recursively composed to express nested scopes (e.g., the final sentence of every paragraph).

\paragraph{Target} $\mathcal{T}_i$ specifies objects measured within the selected \textit{Scope}.
We organize 15 subcategories into six major categories:
\textbf{(1) Capacity} covers text length and unit counts (e.g., sentences or paragraphs), 
\textbf{(2) Literals} tracks specific strings and literal classes (e.g., keywords or punctuation), 
\textbf{(3) Structure} captures containers, members, and nesting depth in structured formats (e.g., JSON or Markdown), 
\textbf{(4) Checklist} assesses mandatory elements and valid values (e.g., specific JSON keys),
\textbf{(5) Pattern} verifies conformity to boundary, skeleton, and surface templates (e.g., enclosed quotes), 
and \textbf{(6) Relations} characterizes multi-object dependencies (e.g., alphabetical ordering or acrostic poetry).

\paragraph{Range} $\mathcal{R}_i$ specifies the admissible values of \textit{Target} through four absolute and two relative categories.
The absolute categories comprise: \textbf{(1) One-Sided Bounds} imposes a minimum or maximum, 
\textbf{(2) Prohibition} fixes the value at zero (e.g., forbidding keywords), 
\textbf{(3) Equality} prescribes an exact value (e.g., exactly five sentences), 
and \textbf{(4) Intervals} defines a closed numeric range (e.g., 80 to 120 words).
The relative categories derive admissible values from another response measurement: 
\textbf{(5) Comparison} imposes inequalities (e.g., increasing word counts per sentence),
and \textbf{(6) Arithmetic Relations} enforces precise equations.
Crucially, this schema accommodates qualitative constraints (e.g., valid JSON format) by formulating compliance as an \textbf{Equality} \textit{Range} of one.

\subsection{Instruction Generation Pipeline}
\paragraph{Atomic Constraint Generation.} 
To gather a diverse and representative pool of atomic constraints, we enumerate all compatible combinations of \textit{Scope}, \textit{Target}, and \textit{Range} categories. 
For each combination, we chain 10 generation iterations.
In each iteration, we prompt a strong LLM to generate a batch of five distinct instructions from a seed, each incorporating an atomic constraint instantiated from that combination.
The seed instruction grounds synthesis in realistic contexts, while batch generation improves efficiency. 
To avoid repetition, previously generated constraints for the same combination are provided as context history, guiding the LLM to elicit novel realizations.

\paragraph{LLM-Based Constraint Crossover.}
After establishing the constraint pool, we perform constraint crossover to synthesize multi-constraint instructions of varying complexity. 
Guided by a predefined target $k\in\{2,4,6,8\}$, we randomly sample 16 candidate constraints per iteration and prompt a strong LLM to construct five distinct instructions from a seed, each integrating at least $k$ of these constraints. 
During integration, the LLM is permitted to modify both the seed and the sampled constraints to ensure fluency and coherence. 
To facilitate downstream verification, the LLM is also required to output a constraint list for each instruction, alongside \textit{Target} for each constraint.

\paragraph{Data Curation and Quality Control.}
We derive seed instructions from HIR-16k \citep{zhang2025replay} and MulDimIF \citep{ye2026muldimif}, applying the above process to construct instructions with single or multiple scope-aware constraints.
Then, we employ strong LLMs to assess the quality of instructions and verify their constraint lists, discarding any instances with ambiguous intents, semantic conflicts, unreasonable requests, or inaccurate constraint lists. 
Finally, we sample the retained instructions across varying constraint counts to ensure diverse instruction complexity.
The resulting \textsc{ScopeInstruct} comprises 16,968 training and 1,000 test instructions, each containing 1 to 12 constraints (4.83 on average).
Detailed data statistics of \textsc{ScopeInstruct} are in Appendix \ref{app:data_statistic}.

%% file: sections/6_methodology.tex
In this section, we introduce ScopeIF, an RL training framework for scope-aware precise instruction-following. 
We first employ a tool-grounded verification pipeline to derive granular response statistics for each constraint. 
A graded reward modeling mechanism is then applied to quantify constraint violations and provide fine-grained reward signals. 
Without loss of generality, we adopt the Group Relative Policy Optimization (GRPO) \citep{shao2024deepseekmath} algorithm for RL training.

\subsection{Tool-Grounded Constraint Verification}
\label{sec:tool_grounded_constraint_verification}
Verifying scope-aware constraints is challenging, as it requires both precisely isolating relevant response segments and counting elements.
For example, verifying \textit{Generate 10 titles, each containing 8 words} requires extracting all titles in the response before calculating length, yet titles can be delimited in diverse ways.
Rule-based verifiers offer precision but are limited to rigid constraints and fixed delimitation, whereas LLM judges are flexible but struggle with exact counting.
To bridge this gap, we empower judge models to dynamically design and execute verification tools conditioned on the response rather than statically pre-defined for constraint alone \citep{peng-etal-2025-agentic, wen2024benchmarking}, thereby accommodating response-specific delimitation.
Specifically, for instruction $x$, its constraint $c_i$ with \textit{Target} $\mathcal{T}_i=\{t_{ij}\}_{j=1}^{M_i}$, and response $y$, our verification pipeline operates as follows:

\paragraph{Tool Design and Execution.} 
To compute any auxiliary evidence required for deriving the statistics of $\mathcal{T}_i$, the judge model $J$ first designs a custom Python tool $f_i$ (if needed) along with its input arguments $\mathbf{z}_i$ tailored for $y$:
\begin{equation}
(f_i,\mathbf{z}_i)
=
J(x,c_i,\mathcal{T}_i,y)
\label{eq:tool_generation}
\end{equation}
An interpreter then executes the generated code to yield this evidence $\mathbf{e}_i$ for accurate verification:
\begin{equation}
\mathbf{e}_i
=
f_i(\mathbf{z}_i)
\label{eq:tool_execution}
\end{equation}
\paragraph{Tool-Grounded Judgment.}
Grounded in this evidence and response content, the judge model then generates granular statistics $\mathcal{V}_i(y)$ for $\mathcal{T}_i$, which serve as the verification result for constraint $c_i$:
\begin{equation}
\mathcal{V}_i(y)
=
J
(x,c_i,\mathcal{T}_i,y,\mathbf{e}_i)
\label{eq:tool_grounded_judgment}
\end{equation}
Specifically, $\mathcal{V}_i(y)$ is a set of quantitative records $\left\{ \left( [l_{ij},u_{ij}], 
a_{ij} \right) \right\}_{j=1}^{M_i}$, 
each pairing the permissible bound $[l_{ij},u_{ij}]$ of object $t_{ij}$ with its measured value $a_{ij}$.
These records yield the binary satisfaction indicator $I_i(y) = \mathbbm{1}(\forall j, a_{ij} \in [l_{ij}, u_{ij}])$ and enable the subsequent graded reward computation.

\subsection{Graded Constraint Reward Modeling}
\label{sec:graded_constraint_reward_modeling}

To alleviate the sparsity of binary rewards for complex constraints, we quantify each constraint's violation degree from its verification statistics and hierarchically aggregate the resulting graded reward with the binary satisfaction signal of each constraint into a denser response-level reward.

\paragraph{Graded Reward Calculation.}
For each constraint $c_i$, we first quantify the violation degree of each object $t_{ij} \in \mathcal{T}_i$.
To normalize violations across
heterogeneous scales, we compute a normalized violation metric $v_{ij} = d_{ij}/s_{ij}$, where $d_{ij}$ is the absolute deviation of the measured value $a_{ij}$ from the permissible bound $[l_{ij}, u_{ij}]$ (with $d_{ij}=0$ if $a_{ij} \in [l_{ij}, u_{ij}]$), and $s_{ij}$ denotes its characteristic range magnitude. 
The graded reward of $t_{ij}$ is then computed via exponential decay:
\begin{equation}
r_{ij} = \exp(-\alpha v_{ij})
\label{eq:target_reward}
\end{equation}
where $\alpha$ controls the penalty severity, and $r_{ij}=1$ indicates full satisfaction.
We then derive the graded reward $g_i$ of $c_i$ as the geometric mean of rewards across all objects in $\mathcal{T}_i$:
\begin{equation}
g_i = \sqrt[M_i]{\prod_{j=1}^{M_i} r_{ij}}
\label{eq:constraint_reward}
\end{equation}

\paragraph{Hierarchical Final Reward Aggregation.}
Directly averaging graded rewards $\{g_i\}$ risks reward misalignment, where accumulating partial scores may outweigh fully satisfying constraints.
To address this problem, we hierarchically aggregate the binary satisfaction indicator $I_i(y)$ with the mean graded reward of the violated constraint subset $\mathcal{F} = \{i \mid I_i(y)=0\}$ (taken as $0$ if $ \mathcal{F} = \varnothing $):
\begin{equation}
R(x,y) = \frac{1}{|C|} \Biggl(
\underbrace{\sum_{i=1}^{|C|} I_i(y)}_{\text{binary reward}}
+ \underbrace{\frac{1}{|\mathcal{F}|} \sum_{i \in \mathcal{F}} g_i}_{\text{graded reward}}
\Biggr)
\label{eq:final_reward}
\end{equation}
This formulation bounds $R(x,y) \in [m/\vert{}C\vert{}, (m+1)/\vert{}C\vert{})$ when $m$ constraints are fully satisfied.
Thus, satisfying more constraints guarantees a discrete reward leap, while the graded reward distinguishes degrees of partial compliance and provides dense supervision for failures. 
Finally, this reward is integrated with a format reward for GRPO training.
More details are in Appendix \ref{app:reward_calculation}.

%% file: sections/7_experiments.tex
\subsection{Experimental Setup}
\paragraph{Evaluation Benchmarks.}
Alongside our test set, we evaluate precise instruction-following on three public benchmarks, including IFEval \citep{zhou2023instruction}, IFBench \citep{pyatkin2026generalizing}, and IFHierBench \citep{mao2026ifhierbench}.
IFEval focuses on simple global objective constraints, while IFBench and IFHierBench focus on more complex scope-aware constraints.
Additionally, we evaluate our method across three general benchmarks, including MATH-500 \citep{lightman2024let}, GPQA \citep{rein2023gpqa}, and MMLU-Pro \citep{wang2024mmlu}. 
Following ImpRIF \citep{yang-etal-2026-imprif}, our test set uses two metrics: CSR (Constraint Success Rate) for the proportion of individual constraints satisfied, and ISR (Instruction Success Rate) for the proportion of samples where all constraints are satisfied.
Details of evaluation metrics are presented in Appendix \ref{app:evaluation_metrics}.
\input{tables/main_results_methods}
\input{tables/main_results_datasets}
\paragraph{Baselines.}
We choose Qwen3-4B \citep{yang2025qwen3} and Qwen3-8B as initial policy models. 
The baselines encompass three categories: 1) Off-the-shelf LLMs, including the original policy models alongside several strong open-source and proprietary LLMs; 2) Alternative optimization methods based on \textsc{ScopeInstruct} and our constraint verification method, including Direct Preference Optimization (DPO) \citep{rafailov2023direct}, Reinforcement Learning (RL) with Instruction-Level Accuracy as the reward \citep{peng-etal-2025-verif}, and RL with Constraint-Level Accuracy as the reward \citep{qin2026incentivizing, liu2025recast}; and 3) Existing instruction-following improving frameworks, including HIR \citep{zhang2025replay}, VerIF \citep{peng-etal-2025-verif}, and MulDimIF \citep{ye2026muldimif}.

\paragraph{Implementation Details.}
For data construction, we use Seed-2.0-Pro \citep{seed2026seed} for constraint generation and crossover, alongside GPT-OSS-120B\footnote{https://huggingface.co/openai/gpt-oss-120b} for instruction quality control.
For reward calculation, the penalty severity $\alpha$ is set to 3.
For GRPO training, we use the verl \citep{sheng2025hybridflow} framework, performing 8 rollouts per instruction with a batch size of 32. 
Across all settings, GPT-OSS-120B serves as the judge model, with its reasoning effort configured to medium.
Greedy search is used for evaluation to ensure reproducibility.
More details are in Appendix \ref{app:training_details}.

\subsection{Main Results}
Tables \ref{tab:main_result_methods} and \ref{tab:main_result_datasets} present the performance of different optimization methods and instruction-following frameworks.
\textbf{Firstly}, ScopeIF significantly enhances scope-aware instruction-following capabilities across different model scales.
With average relative improvements of 60.7\% and 44.7\% on IFBench and IFHierBench, the optimized Qwen3-4B and 8B models can rival or even surpass strong frontier models such as Gemini-2.5-Pro and DeepSeek-V3.2.
\textbf{Secondly}, ScopeIF outperforms all alternative optimization methods. 
RL approaches substantially beat DPO, and their effectiveness strongly correlates with reward granularity. 
Average performance improves progressively as rewards transition from ILA to CLA and peaks with ScopeIF, which validates the value of our fine-grained graded constraint reward modeling mechanism.
We also conduct an ablation study that removes hierarchical reward aggregation by directly averaging graded rewards.
Results in Table \ref{tab:main_result_methods} show performance drops on most metrics, highlighting the importance of hierarchical aggregation in encouraging full constraint compliance.
\textbf{Thirdly}, ScopeIF consistently surpasses existing instruction-following improving frameworks, including HIR, VerIF, and MulDimIF. 
While the performance gap is marginal on IFEval's simple constraints, ScopeIF establishes a substantial lead on complex scope-aware constraints across the other three benchmarks. 
This highlights the oversight of scope-aware constraints in prior methods and the importance of our unified schema for constraint scope modeling.
\textbf{Finally}, as shown in Table \ref{tab:general_performance}, ScopeIF preserves the general performance of LLMs with negligible degradation.

\begin{wrapfigure}{r}{0.48\linewidth}
  \vspace{-4mm}
  \centering
  \includegraphics[width=1.0\linewidth]{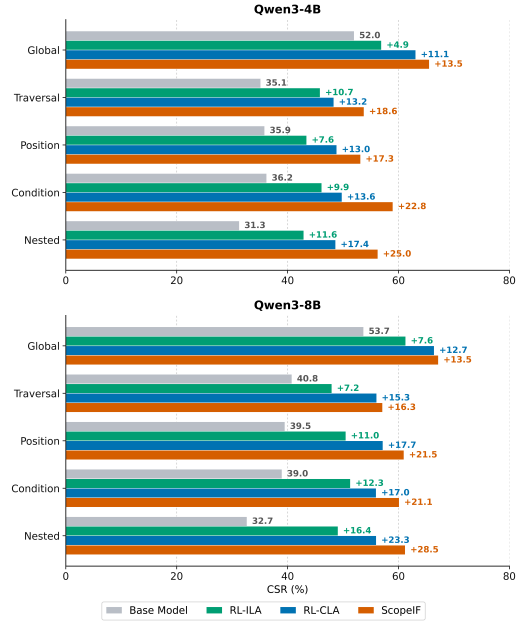}
  \vspace{-6mm}
  \caption{Performance of constraints with different scope categories on our test set.}
  \label{fig:performance_per_scope}
  \vspace{-6mm}
\end{wrapfigure}

\subsection{Analysis}
\paragraph{Performance on Different Constraint Scopes.}
To dissect the ability of LLMs to follow constraints with specific scopes, we report the CSR of each \textit{Scope} category on our test set in Figure \ref{fig:performance_per_scope}.
\textbf{Firstly}, scope complexity poses a growing challenge to LLMs.
While base models perform best on Global, their performance drops markedly on single-layer scope selectors (i.e., Traversal, Position, and Condition), and reaches its lowest point on recursively composed Nested scopes. 
This trend highlights the inherent difficulty of scope-aware instruction-following and aligns with our motivation.
\textbf{Secondly}, the advantage of ScopeIF over binary-reward baselines widens as scope complexity increases, which is marginal on Global yet grows steadily toward single-layer and nested scopes.
Complex scopes render binary rewards increasingly sparse, whereas our graded reward still quantifies partial violations to yield denser learning signals, confirming its particular value for complex scope-aware constraints.
We also analyze the performance of LLMs on different constraint \textit{Target} and \textit{Range} categories in Appendix \ref{app:detailed_results}.

\begin{table}[!t]
\centering
\newsavebox{\scopeifGeneralTableBox}
\newsavebox{\scopeifJudgeTableBox}
\sbox{\scopeifGeneralTableBox}{\input{tables/general_performance}\unskip}
\sbox{\scopeifJudgeTableBox}{\input{tables/judge_models_performance}\unskip}
\begin{minipage}[t][\dimexpr\ht\scopeifJudgeTableBox+\dp\scopeifJudgeTableBox\relax][s]{0.435\linewidth}
\vspace{0pt}
\usebox{\scopeifGeneralTableBox}
\par\vfill
\input{tables/constraint_diversity}\par\vspace{0pt}
\end{minipage}\hfill
\usebox{\scopeifJudgeTableBox}
% \vspace{-4mm}
\end{table}

\paragraph{Constraint Verification Methods.} 
To validate the reliability of our verification methods, we construct a meta-evaluation set of 500 instructions sampled from our test set, each paired with responses from 2 of 6 LLMs of varying capability.
Two annotators independently judge compliance for each constraint, with a third inspector cross-validating and resolving discrepancies by consensus.
The resulting dataset comprises 5,350 constraints, of which 1,843 are labeled as violated.
Under three judge models, we compare our Tool-Grounded Verification (TGV) method with pure LLM judges (LLM) and rule-based verifiers (Rule) using two metrics: pointwise accuracy, the per-constraint agreement with human labels; and pairwise agreement, the fraction of response pairs for which the judge and humans agree on which response satisfies more constraints.
As shown in Table \ref{tab:judge_models_performance}, TGV combines the flexibility of LLM judges with the precision of rule-based verifiers through dynamic tool generation, substantially outperforming both baselines across all judge models.
Moreover, generating the verification tool without conditioning on the response (w/ Static Tool) causes a consistent drop, confirming the benefit of tailoring custom tools to each response.
Beyond our own test set, we further evaluate these methods under the constraint assessment setting of IF-RewardBench \citep{wen-etal-2026-rewardbench}, which measures the ability of verifiers to rank the instruction-following quality of multiple responses.
Here, TGV also achieves the best performance, confirming its generalization to diverse constraint types.
More details are in Appendix \ref{app:constraint_verification}.

\begin{figure}[!t]
  \centering
  % \vspace{-2mm}
  \includegraphics[width=1.0\linewidth]{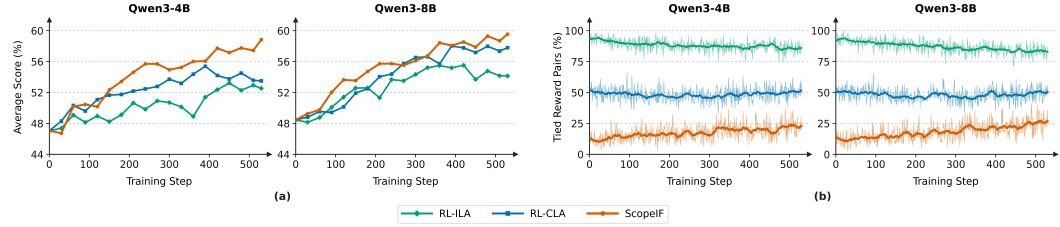}
  % \vspace{-6mm}
  \caption{Average performance of IFEval, IFBench, and IFHierBench (a), and the proportion of rollout response pairs with identical rewards (b) during ScopeIF training.}
  \label{fig:training_curves}
  % \vspace{-2mm}
  % Reduce the gap between Figure 4 and the following text.
  \vspace{-3mm}
\end{figure}
\paragraph{Constraint Diversity.}
To validate the effectiveness of our factorized constraint synthesis method in improving constraint diversity, we compare the constraint-level diversity of \textsc{ScopeInstruct} with three widely used baseline training datasets.
Three automatic metrics are utilized:
(1) Self-BLEU-4 \citep{zhu2018texygen} measures inter-constraint similarity by treating each constraint as the hypothesis and the remaining constraints as references;
(2) Distinct-2 \citep{li2016diversity} calculates the proportion of unique bigrams among all bigram occurrences; and (3) Entropy-4 \citep{zhang2018generating} quantifies the diversity and distributional balance of four-grams using Shannon entropy.
Table \ref{tab:constraint_diversity} shows that \textsc{ScopeInstruct} consistently exhibits greater diversity across these measures, supporting the benefit of factorizing and recombining different dimensions for diverse constraint synthesis. 

\paragraph{Training Curves.}
To examine the training dynamics of different reward designs, we trace the average score on IFEval, IFBench, and IFHierBench throughout RL training in Figure \ref{fig:training_curves} (a).
\textbf{Firstly}, ScopeIF outperforms both binary-reward baselines throughout the training process, demonstrating a robust and consistent advantage.
\textbf{Secondly}, this advantage is more pronounced on Qwen3-4B than on Qwen3-8B, as smaller models may be more prone to the sparse-reward problem and thus benefit more from the denser signals of our graded reward.
\textbf{Thirdly}, the curves have not yet plateaued under our current budget, with ScopeIF still trending upward at the final step.
This suggests that further scaling of training resources, such as more training steps or rollouts, could unlock additional gains.
To quantify the density of different reward mechanisms, we also trace the proportion of response pairs receiving identical rewards within each GRPO rollout group throughout RL training in Figure \ref{fig:training_curves} (b), 
as identical rewards provide no relative preference between the paired responses.
% which provides no preference signal under the group-wise advantage normalization of GRPO.
This ratio drops sharply as reward granularity increases from ILA to CLA and ScopeIF, and the gap persists throughout training, showing that our graded reward substantially alleviates reward sparsity.

%% file: tables/main_results_methods.tex
\begin{table*}[!t]
% \centering % Replaced by the official center environment below.
\small
\caption{Experimental results (\%) of different optimization methods on instruction-following benchmarks. 
P and I stand for prompt-level and instruction-level metrics, respectively. 
L and S denote loose and strict evaluations, respectively. HRA means hierarchical reward aggregation.
The highest result for each backbone model is \textbf{bolded}, and the overall highest is \underline{underlined}. }
\label{tab:main_result_methods}
\begin{center}
\resizebox{\textwidth}{!} {
\begin{tabular}{l|cc|cccc|cccc|cc|c}
\toprule
\multirow{2}{*}{\textbf{Model}} & \multicolumn{2}{c|}{\textbf{ScopeIF-Test}} & \multicolumn{4}{c|}{\textbf{IFEval}} & \multicolumn{4}{c|}{\textbf{IFBench}} & \multicolumn{2}{c|}{\textbf{IFHierBench}} & \multirow{2}{*}{\textbf{Avg.}} \\
\cmidrule{2-13}
& ISR & CSR & P-S & P-L & I-S & I-L & P-S & P-L & I-S & I-L & P-S & I-S & \\
\midrule
\multicolumn{14}{c}{\textit{Frontier Models}} \\
\midrule
Seed-2.0-Pro & \underline{21.8} & \underline{69.5} & 88.2 & 91.7 & 92.3 & 94.6 & 49.3 & 56.0 & 51.5 & 57.6 & \underline{22.8} & \underline{52.8} & \underline{57.2} \\
Gemini-2.5-Pro & 15.8 & 63.7 & 90.0 & \underline{93.0} & 92.8 & \underline{95.2} & 40.0 & 47.7 & 43.0 & 50.9 & 20.8 & 46.7 & 52.9 \\
DeepSeek-V3.2 & 9.1  & 50.8 & 87.4 & 91.5 & 91.2 & 94.0 & 38.3 & 43.3 & 42.2 & 47.1 & 14.3 & 33.7 & 46.9 \\
GPT-4.1 & 8.6 & 53.1 & 87.8 & 91.3 & 91.5 & 94.4 & 36.7 & 41.3 & 41.3 & 46.2 & 18.2 & 44.1 & 48.7 \\
QwQ-32B & 8.5 & 48.7 & 82.4 & 87.6 & 88.1 & 91.6 & 34.7 & 39.7 & 38.4 & 43.3 & 17.2 & 37.2 & 45.6 \\
\midrule
\multicolumn{14}{c}{\textit{Our Models}} \\
\midrule
Qwen3-4B & 5.2 & 40.6 & 81.7 & 85.6 & 86.8 & 89.6 & 29.7 & 34.3 & 32.8 & 38.1 & 15.0 & 25.0 & 40.6 \\
\quad + DPO & 5.6 & 43.5 & 86.1 & 88.7 & 90.6 & 92.6 & 32.0 & 36.7 & 34.6 & 40.7 & 15.7 & 27.6 & 42.9 \\
\quad + RL-ILA & 10.5 & 50.5 & 85.4 & 87.4 & 89.9 & 91.5 & 47.7 & 50.0 & 50.9 & 53.8 & 17.3 & 22.0 & 47.3 \\
\quad + RL-CLA & 12.8 & 55.2 & 85.8 & 89.6 & 90.4 & 92.9 & 48.0 & 51.3 & 50.3 & 53.5 & 18.8 & 32.6 & 50.0 \\
\rowcolor{rowblue} \quad + ScopeIF (Ours) & \textbf{15.3} & \textbf{60.9} & \textbf{88.2} & \textbf{90.9} & \textbf{92.2} & \textbf{94.1} & \textbf{49.7} & \textbf{53.3} & \textbf{52.0} & \textbf{56.1} & 18.8 & \textbf{39.4} & \textbf{52.8} \\
\quad \quad \textit{- wo HRA} & 14.1 & 58.6 & 87.2 & 90.8 & 91.6 & 93.9 & 48.0 & 50.7 & 50.0 & 52.6 & \textbf{20.7} & 36.7 & 51.6 \\
\midrule
Qwen3-8B & 7.2 & 43.8 & 85.0 & 88.4 & 89.3 & 91.6  & 31.0 & 33.7 & 34.9 & 38.4 & 15.0 & 25.5 & 42.2 \\
\quad + DPO & 6.8 & 45.2 & 85.6 & 89.3 & 90.0 & 92.8 & 33.7 & 39.3 & 36.3 & 42.4 & 15.8 & 26.7 & 43.7 \\
\quad + RL-ILA & 14.2 & 55.1 & 87.1 & 88.5 & 91.1 & 92.2 & 45.7 & 50.0 & 47.0 & 51.5 & 18.2 & 31.3 & 49.4 \\
\quad + RL-CLA & 17.7 & 61.9 & 88.4 & 91.1 & 92.2 & 94.0 & 51.0 & 55.0 & 52.9 & 56.7 & 20.5 & 35.6 & 53.3 \\
\rowcolor{rowblue} \quad + ScopeIF (Ours) & \textbf{19.6} & \textbf{64.9} & \textbf{\underline{90.4}} & \textbf{92.2} & \textbf{\underline{93.3}} & \textbf{94.6} & \textbf{\underline{53.7}} & \textbf{\underline{57.0}} & \textbf{\underline{57.0}} & \textbf{\underline{59.9}} & \textbf{21.7} & 36.6 & \textbf{55.2} \\
\quad \quad \textit{- wo HRA} & 16.0 & 62.2 & 88.4 & 90.8 & 91.8 & 93.5 & 50.3 & 55.0 & 52.9 & 57.3 & 21.3 & \textbf{38.6} & 53.5 \\
\bottomrule
\end{tabular}
}
\end{center}
% Reduce the space below this table in the arXiv version.
\vspace{-6mm}
% Original bottom caption retained; the active caption is above the table.
% \caption{Experimental results (\%) of different optimization methods on instruction-following benchmarks. 
% P and I stand for prompt-level and instruction-level metrics, respectively. 
% L and S denote loose and strict evaluations, respectively. HRA means hierarchical reward aggregation.
% The highest result for each backbone model is \textbf{bolded}, and the overall highest is \underline{underlined}. }
% \label{tab:main_result_methods}
\end{table*}

%% file: tables/main_results_datasets.tex
\begin{table*}[!t]
% \centering % Replaced by the official center environment below.
\small
\caption{Experimental results (\%) of different instruction-following improving frameworks. To ensure fairness in the amount of training data, MulDimIF used twice the number of training epochs.}
\label{tab:main_result_datasets}
\begin{center}
% Match Table 1 by raising the table body without moving its caption.
\vspace{-2.75mm}
\resizebox{\textwidth}{!} {
\begin{tabular}{llc|cc|cccc|cccc|cc|c}
\toprule
\multirow{2}{*}{\textbf{Model}} & \multirow{2}{*}{\textbf{Method}} & \multirow{2}{*}{\textbf{\#Data}} & \multicolumn{2}{c|}{\textbf{ScopeIF-Test}} & \multicolumn{4}{c|}{\textbf{IFEval}} & \multicolumn{4}{c|}{\textbf{IFBench}} & \multicolumn{2}{c|}{\textbf{IFHierBench}} & \multirow{2}{*}{\textbf{Avg.}} \\
\cmidrule{4-15}
& & & ISR & CSR & P-S & P-L & I-S & I-L & P-S & P-L & I-S & I-L & P-S & I-S & \\
\midrule
 & HIR & 16k & 6.7 & 44.2 & 87.2 & \textbf{91.0} & 90.4 & 92.9 & 41.7 & 44.7 & 44.2 & 47.7 & 16.0 & 25.4 & 45.3 \\
 & VerIF & 22k & 7.9 & 47.7 & 86.5 & 89.8 & 91.0 & 93.3 & 42.3 & 48.3 & 45.3 & 50.9 & 16.5 & 30.9 & 47.1 \\
 & MulDimIF & 8k & 6.0 & 42.6 & 85.6 & 88.5 & 90.4 & 92.3 & 39.3 & 43.0 & 41.6 & 46.5 & 14.7 & 21.6 & 43.6 \\
\rowcolor{rowblue} \cellcolor{white}\multirow{-4}{*}{Qwen3-4B} & ScopeIF (Ours) & 16k & \textbf{15.3} & \textbf{60.9} & \textbf{88.2} & 90.9 & \textbf{92.2} & \textbf{94.1} & \textbf{49.7} & \textbf{53.3} & \textbf{52.0} & \textbf{56.1} & \textbf{18.8} & \textbf{39.4} & \textbf{52.8} \\
\midrule
 & HIR & 16k & 8.6 & 48.9 & 89.5 & 92.1 & 92.7 & 94.6 & 39.3 & 44.3 & 41.9 & 46.8 & 17.3 & 30.7 & 47.0 \\
 & VerIF & 22k & 10.0 & 46.9 & 87.1 & 91.1 & 91.1 & 94.0 & 42.0 & 47.3 & 44.8 & 50.3 & 16.7 & 27.9 & 46.9 \\
 & MulDimIF & 8k & 6.4 & 43.5 & 86.7 & 89.5 & 91.2 & 93.0 & 34.7 & 40.7 & 38.1 & 43.3 & 15.8 & 20.5 & 43.1 \\
\rowcolor{rowblue} \cellcolor{white}\multirow{-4}{*}{Qwen3-8B} & ScopeIF (Ours) & 16k & \textbf{19.6}  & \textbf{64.9} & \textbf{90.4} & \textbf{92.2} & \textbf{93.3} & \textbf{94.6} & \textbf{53.7} & \textbf{57.0} & \textbf{57.0} & \textbf{59.9} & \textbf{21.7} & \textbf{36.6} & \textbf{55.2} \\
\bottomrule
\end{tabular}
}
\end{center}
% Reduce the space below this table in the arXiv version.
% \vspace{-6mm}
% Retain the space freed by raising the table body below the table.
\vspace{-3.25mm}
% Original bottom caption retained; the active caption is above the table.
% \caption{Experimental results (\%) of different instruction-following improving frameworks. To ensure fairness in the amount of training data, MulDimIF used twice the number of training epochs.}
% \vspace{-4mm}
% \label{tab:main_result_datasets}
\end{table*}

%% file: tables/general_performance.tex
\begin{minipage}[t]{0.435\linewidth}
\vspace{0pt}
% Cancel the caption leading space at the top of this table block.
\vspace{-\abovecaptionskip}
% \centering % Replaced by the official center environment below.
\captionof{table}{Performance (\%) of ScopeIF on general benchmarks.}
\label{tab:general_performance}
\begin{center}
% Move the table body closer to its caption; retain the caption alignment.
\vspace{-3mm}
{\small
\resizebox{\linewidth}{!}{%
\begin{tabular}{l|ccc}
\toprule
\textbf{Model} & \textbf{MATH-500} & \textbf{GPQA} & \textbf{MMLU-Pro} \\
\midrule
Qwen3-4B & 87.8 & 54.0 & 70.7 \\
\quad + ScopeIF (Ours) & 87.6 \textcolor[RGB]{204,0,0}{$\downarrow$0.2} & 53.0 \textcolor[RGB]{204,0,0}{$\downarrow$1.0} & 70.0 \textcolor[RGB]{204,0,0}{$\downarrow$0.7} \\
\midrule
Qwen3-8B & 89.2 & 59.6 & 74.8 \\
\quad + ScopeIF (Ours) & 88.6 \textcolor[RGB]{204,0,0}{$\downarrow$0.6} & 60.1 \textcolor[RGB]{0,153,0}{$\uparrow$0.5} & 75.2 \textcolor[RGB]{0,153,0}{$\uparrow$0.4} \\
\bottomrule
\end{tabular}}}
\end{center}
% Original bottom caption retained; the active caption is above the table.
% \captionof{table}{Performance (\%) of ScopeIF on general benchmarks.}
% \label{tab:general_performance}
\end{minipage}

%% file: tables/judge_models_performance.tex
\begin{minipage}[t]{0.530\linewidth}
\vspace{0pt}
% Cancel the caption leading space at the top of this table block.
\vspace{-\abovecaptionskip}
% \centering % Replaced by the official center environment below.
\caption{Pointwise accuracy (Acc.) and pairwise agreement (Agr.) of judge models on our test set and Kendall correlation $\tau$ on IF-RewardBench under the constraint assessment setting.}
\label{tab:judge_models_performance}
\begin{center}
{\small
\resizebox{\linewidth}{!}{%
\begin{tabular}{ll|cc|c}
\toprule
\multirow{2}{*}{\textbf{Judge Model}} & \multirow{2}{*}{\textbf{Method}} & \multicolumn{2}{c|}{\textbf{ScopeIF-Test}} & \textbf{IF-RewardBench} \\
\cmidrule{3-5}
& & Acc. & Agr. & $\tau$ \\
\midrule
\multirow{4}{*}{Qwen3-32B} & LLM & 0.820 & 0.606 & 0.351 \\
 & Rule & 0.799 & 0.668 & 0.372 \\
 & TGV (Ours) & \textbf{0.881} & \textbf{0.734} & \textbf{0.461} \\
 & - \textit{w/ Static Tool} & 0.864 & 0.694 & 0.457 \\
\midrule
\multirow{4}{*}{QwQ-32B} & LLM & 0.867 & 0.712 & 0.494 \\
 & Rule & 0.841 & 0.726 & 0.419 \\
 & TGV (Ours) & \textbf{0.902} & \textbf{0.792} & \textbf{0.532} \\
 & - \textit{w/ Static Tool} & 0.894 & 0.776 & 0.513 \\
\midrule
\multirow{4}{*}{GPT-OSS-120B} & LLM & 0.923 & 0.810 & 0.512 \\
 & Rule & 0.860 & 0.738 & 0.388 \\
 & TGV (Ours) & \textbf{0.935} & \textbf{0.842} & \textbf{0.558} \\
 & - \textit{w/ Static Tool} & 0.922 & 0.804 & 0.548 \\
\bottomrule
\end{tabular}}}
\end{center}
% Original bottom caption retained; the active caption is above the table.
% \caption{Pointwise accuracy (Acc.) and pairwise agreement (Agr.) of judge models on our test set and Kendall correlation $\tau$ on IF-RewardBench under the constraint assessment setting.}
% \label{tab:judge_models_performance}
\end{minipage}

%% file: tables/constraint_diversity.tex
\begin{minipage}[t]{\linewidth}
    % \centering % Replaced by the official center environment below.
    \small
    \caption{Constraint diversity of different training datasets for instruction-following.}
\label{tab:constraint_diversity}
\begin{center}
% Tighten this gap; the outer bottom alignment moves the caption down.
\vspace{-2.75mm}
\resizebox{\linewidth}{!}{%
    \begin{tabular}{lccc}
        \toprule
        \textbf{Dataset} & \textbf{Self-BLEU-4} $\boldsymbol{\downarrow}$ & \textbf{Distinct-2} $\boldsymbol{\uparrow}$ & \textbf{Entropy-4} $\boldsymbol{\uparrow}$ \\
        \midrule
        HIR       & 0.931 & 0.067 & 10.3 \\
        VerIF     & 0.918 & 0.068 & 11.0 \\
        MulDimIF  & 0.961 & 0.059 & 7.8 \\
        \midrule
        \textsc{ScopeInstruct}
                  & \textbf{0.834}
                  & \textbf{0.133}
                  & \textbf{12.8} \\
        \bottomrule
    \end{tabular}}
\end{center}
% Original bottom caption retained; the active caption is above the table.
%     \caption{Constraint diversity of different training datasets for instruction-following.}
%     \label{tab:constraint_diversity}
\end{minipage}

%% file: sections/8_conclusion.tex
We propose ScopeIF, a novel RL training framework for scope-aware precise instruction-following.
We first introduce a unified schema that factorizes objective constraints into three dimensions, \textit{Scope}, \textit{Target}, and \textit{Range}, enabling diverse constraint synthesis and violation degree computation.
Grounded in this, we construct \textsc{ScopeInstruct}, a large-scale dataset of instructions with scope-aware constraints, and combine tool-grounded verification with graded reward modeling for reliable, dense RL supervision.
Extensive experiments show that ScopeIF consistently outperforms existing frameworks and reward modeling paradigms, especially on complex scope-aware constraints.

%% file: sections/9_appendix.tex
\section{Details of Constraint Schema}
\label{app:schema}
In this section, we provide the detailed taxonomy of the three dimensions defined in § \ref{sec:constraint_schema}: \textit{Scope}, \textit{Target}, and
\textit{Range}.

\subsection{Scope}

The \textit{Scope} selector $\mathcal{S}_i$ isolates governed segments by applying a selection operation to response units.
Therefore, the \textit{Scope} of a constraint is instantiated by some response units and a selection operation.

\paragraph{Response Units.}
The selector operates over response units at the following granularities:
\begin{enumerate}
    \item \textbf{Entire Response.}
    The complete response is treated as a single governed segment.

    \item \textbf{Structural Blocks.}
    Complete structural regions, such as sections, lists, tables, and JSON, Markdown, or \LaTeX{} blocks.

    \item \textbf{Structural Elements.}
    Constituent elements within structural blocks, such as list items, table rows or cells, JSON key-value pairs, and Markdown headings.

    \item \textbf{Paragraphs.}
    Individual paragraphs in the response.

    \item \textbf{Sentences.}
    Individual sentences in the response.

    \item \textbf{Words.}
    Individual words in the response.

    \item \textbf{Characters.}
    Individual characters in the response.
\end{enumerate}

\paragraph{Selection Operation.}
Given response units, the selector falls into four categories:
\begin{enumerate}
    \item \textbf{Global} selects the entire response as a governed segment.
    For example, \textit{The response must contain exactly three paragraphs} applies the constraint to the entire response.

    \item \textbf{Traversal} selects every unit of a given kind.
    For example, \textit{Each paragraph must contain at most three sentences} selects every paragraph.

    \item \textbf{Position} selects units at specific locations.
    For example, \textit{The first sentence must contain the keyword ``summary''} targets a sentence by its position.

    \item \textbf{Condition} selects units satisfying a predicate.
    For example, \textit{Sentences containing ``however'' must contain at most 20 words} filters sentences by keyword occurrence.
\end{enumerate}

Selection operations can be recursively composed to express nested scopes.
For example, \textit{The final sentence of every paragraph} combines \textbf{Traversal} over paragraphs with \textbf{Position} within each selected paragraph.

\subsection{Target}

The \textit{Target} $\mathcal{T}_i$ specifies objects measured within the selected \textit{Scope}.
We organize 15 fine-grained subcategories into six major categories:

\begin{enumerate}
    \item \textbf{Capacity} measures text length and natural-language units.
    \begin{enumerate}
        \item \textbf{Text Length.} The number of words or characters.

        \item \textbf{Unit Count.} The number of natural-language units, such as paragraphs, sentences, or lines.
    \end{enumerate}

    \item \textbf{Literals} track occurrences of specific strings or literal classes.
    \begin{enumerate}
        \item \textbf{Specific Strings.} Occurrences of exact characters, words, phrases, or sentences, such as the keyword ``apple''.

        \item \textbf{Literal Classes.} Occurrences of 
        characters or words belonging to a class, such as digits, uppercase letters, emoji, and punctuation marks.
    \end{enumerate}

    \item \textbf{Structure} captures structured containers such as JSON or Markdown, their constituent members, and nesting depth.
    \begin{enumerate}
        \item \textbf{Containers.} Complete structured objects, such as JSON objects, Markdown lists, and CSV tables.

        \item \textbf{Members.} Constituent elements within a structured object, such as JSON key-value pairs, table rows, and list items.

        \item \textbf{Nesting Depth.} The hierarchical depth of a structured object.
    \end{enumerate}

    \item \textbf{Checklist} assesses the inclusion of mandatory elements and valid values.
    \begin{enumerate}
        \item \textbf{Presence.} Inclusion of required components, such as JSON keys \texttt{type}, \texttt{time}, and \texttt{country}.

        \item \textbf{Validity.} Fields or values satisfying predefined specifications, such as an integer \texttt{count} field or a decimal \texttt{price} field.
    \end{enumerate}

    \item \textbf{Pattern} verifies conformity to boundary, skeleton, and surface templates.
    \begin{enumerate}
        \item \textbf{Boundary.} Objects delimited by defined separators, such as text enclosed in quotes or paragraphs separated by ***.

        \item \textbf{Skeleton.} Strings matching a fixed-text skeleton with variable slots, such as \texttt{Question N: ... Answer: ...}.

        \item \textbf{Surface.} Text spans matching a specific format, such as uppercase words or numbers with three decimal places.
    \end{enumerate}

    \item \textbf{Relations} characterize multi-object dependencies.
    \begin{enumerate}
        \item \textbf{Ordering Relations.} Sequential arrangement of objects, such as names sorted alphabetically or events arranged chronologically.

        \item \textbf{Chaining Relations.} Dependencies between adjacent elements, such as the last word of one sentence matching the first word of the next.

        \item \textbf{Positional Relations.} Dependencies across designated positions, such as an acrostic formed by the first letters of consecutive lines.
    \end{enumerate}
\end{enumerate}

\subsection{Range}

The \textit{Range} $\mathcal{R}_i$ specifies the admissible values of \textit{Target}.
It comprises six categories: the first four are absolute, and the following two are relative:

\begin{enumerate}
    \item \textbf{One-Sided Bounds.}
    Impose a minimum or maximum.
    For example, a response should contain at most 100 words.

    \item \textbf{Prohibition.}
    Fix the measured value at zero.
    For example, a response must not contain punctuation marks.

    \item \textbf{Equality.}
    Prescribe an exact value.
    For example, a response should contain exactly five sentences.

    \item \textbf{Intervals.}
    Define a fixed closed numeric range.
    For example, a response should contain between 80 and 120 words.

    \item \textbf{Comparison.}
    Impose an inequality relative to another response measurement.
    For example, each paragraph should contain more words than the preceding paragraph.

    \item \textbf{Arithmetic Relations.}
    Enforce a precise equation relative to another response measurement.
    For example, the third paragraph should contain exactly twice as many words as the second paragraph.
\end{enumerate}

This formulation also accommodates qualitative constraints.
For example, constraints \textit{The response should be a valid JSON object} or \textit{The response should end with ``happy birthday"} can be represented by defining a binary compliance \textit{Target} with an \textbf{Equality} \textit{Range} of one.

\section{List of Prompt Templates}
\label{app:prompt_template}

This section lists all the prompt templates applied
throughout this work, including the prompt for atomic constraint generation in Table \ref{tab:atomic_constraint_generation}, 
the prompt for constraint crossover in Table \ref{tab:constraint_crossover}, 
the prompt for instruction quality assessment in Table \ref{tab:instruction_quality_assessment}, 
the prompt for constraint checklist quality assessment in Table \ref{tab:constraint_checklist_quality_assessment}, the prompt for target extraction for each constraint in Table \ref{tab:target_extraction}, 
the prompt for target extraction quality assessment in Table \ref{tab:target_extraction_quality_assessment}, 
the prompt for tool design in Table \ref{tab:tool_design}, 
and the prompt for tool-grounded judgment in Table \ref{tab:tool_grounded_judgement}.

\begin{figure}[t]
    \centering
    \includegraphics[width=\linewidth]{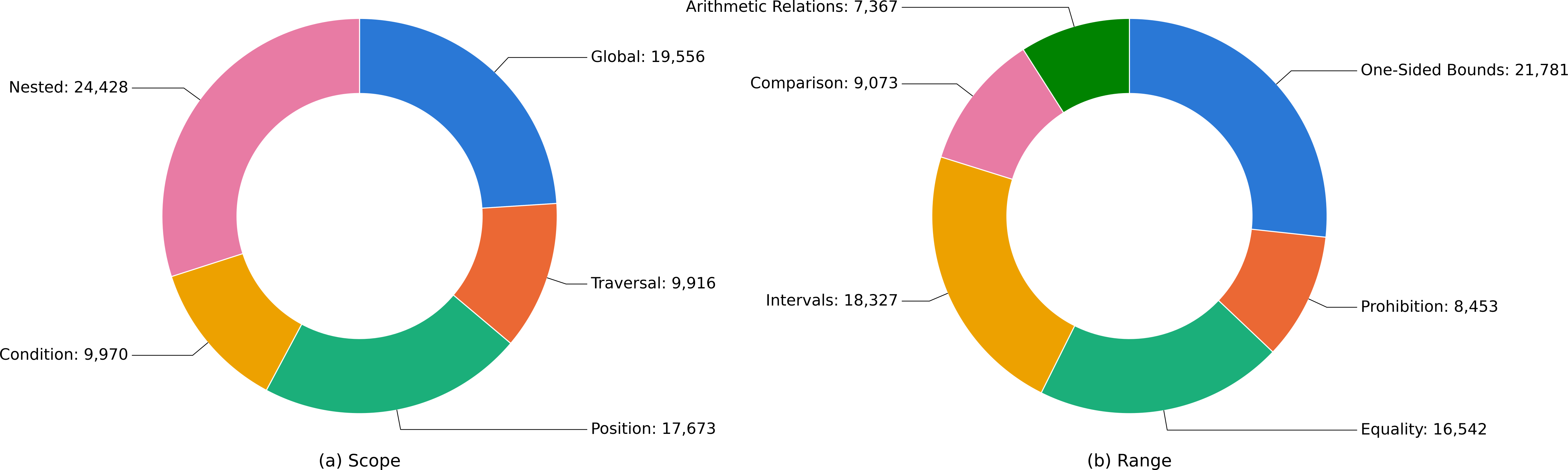}
    \caption{The \textit{Scope} (a) and \textit{Range} (b) distribution of constraints in \textsc{ScopeInstruct}.} \label{fig:constraint_scope_range_distribution}
\end{figure}

\begin{figure}[t]
    \centering
    \includegraphics[width=0.6\linewidth]{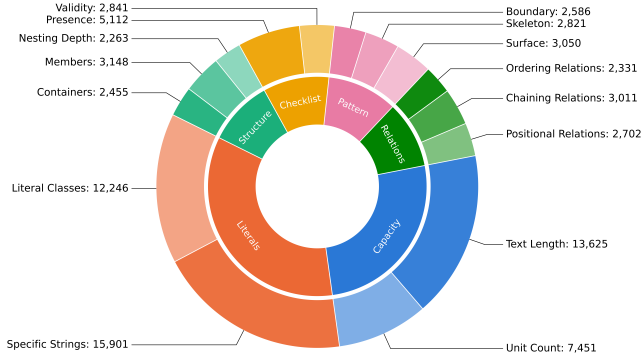}
    \caption{The \textit{Target} distribution of constraints in \textsc{ScopeInstruct}.}
    \label{fig:constraint_target_distribution}
\end{figure}

\section{Data Statistics of \textsc{ScopeInstruct}}
\label{app:data_statistic}
\begin{wrapfigure}{R}{0.48\linewidth}
    \centering
    % \vspace{-4mm} % Original figure offset retained.
    \vspace{-3mm} % Move Figure 7 down by 1 mm.
    \includegraphics[width=\linewidth]{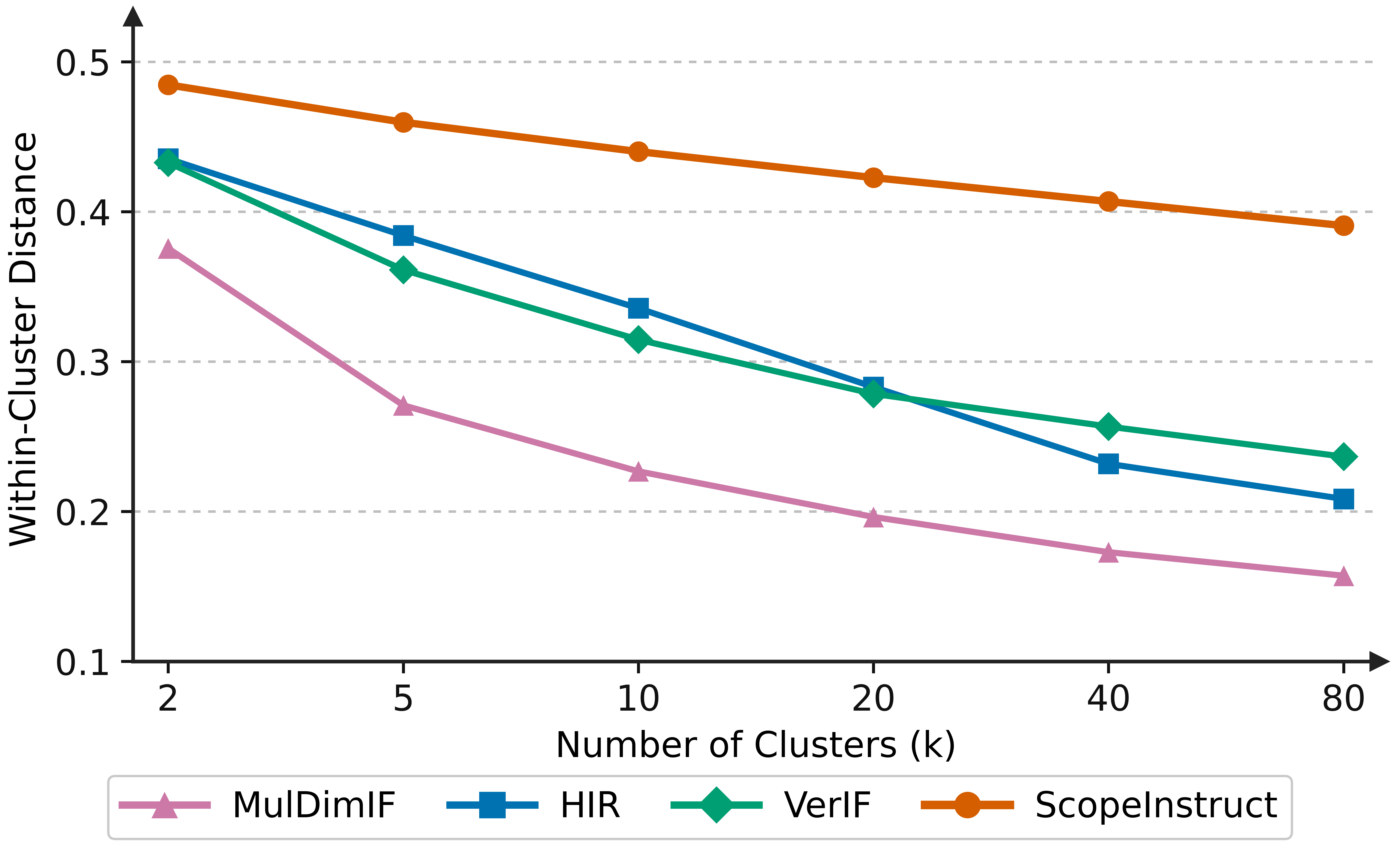}
    % \vspace{-6mm}
    \caption{Mean squared distance from each constraint embedding to its assigned $k$-means centroid under varying numbers of clusters $k$.}
    % \vspace{-2mm}
\label{fig:constraint_diversity_kmeans}
% Keep the original wrap height after shifting the figure down.
\vspace{-1mm}
\end{wrapfigure}
\begin{figure}[t]
    \centering
    \includegraphics[width=0.6\linewidth]{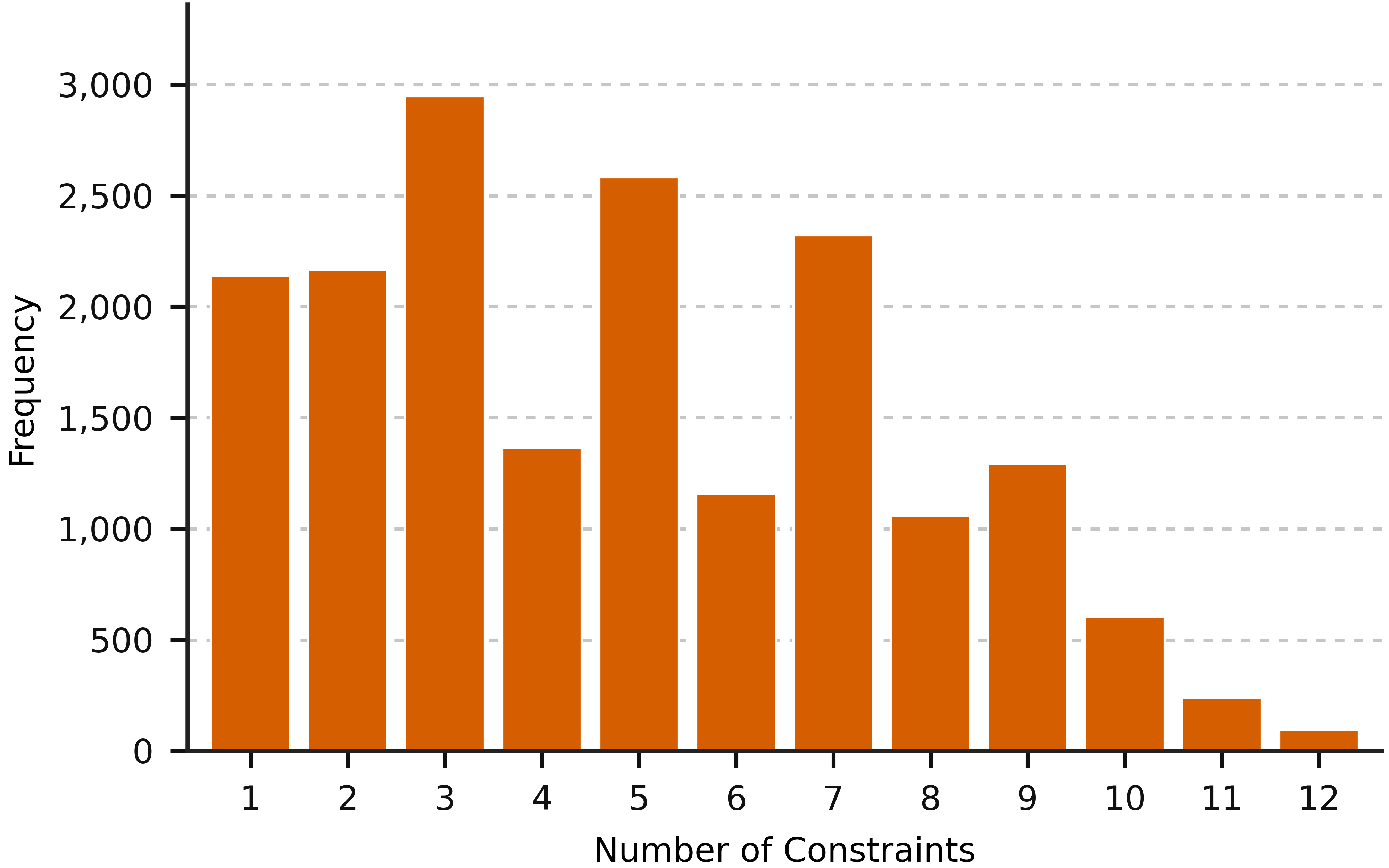}
    \captionof{figure}{Distribution of the number of constraints per instruction in \textsc{ScopeInstruct}.}
    \label{fig:constraint_count_distribution}
\end{figure}

The constructed dataset, \textsc{ScopeInstruct}, contains 17,968 instructions, of which 16,968 are used for training and 1,000 for testing. 
The number of constraints per instruction ranges from 1 to 12, with an average of 4.83. 
Figure \ref{fig:constraint_count_distribution} shows the distribution of constraint counts. Grounded in our constraint schema, \textsc{ScopeInstruct} covers diverse constraint categories across \textit{Scope}, \textit{Target}, and \textit{Range} dimensions, whose distributions are presented in Figures \ref{fig:constraint_scope_range_distribution} and \ref{fig:constraint_target_distribution}.

As the results in Table \ref{tab:constraint_diversity} focus on $n$-gram variations, we further evaluate constraint diversity at the semantic level. Specifically, we extract constraint embeddings using BGE-M3\footnote{https://huggingface.co/BAAI/bge-m3} and apply $k$-means clustering to each dataset separately.
We report the within-cluster distance, i.e., the mean squared distance from each constraint to its assigned centroid, where a lower value indicates that constraints are more templated and reducible to a limited number of prototypes.
As shown in Figure \ref{fig:constraint_diversity_kmeans}, \textsc{ScopeInstruct} attains the highest within-cluster distance at every $k$ and decays far more slowly.
This suggests that constraints of baseline datasets collapse onto a limited set of recurring templates, while our factorized sampling over \textit{Scope}, \textit{Target}, and \textit{Range} produces semantically dispersed combinations even under fine-grained partitioning.

\section{Details of Reward Calculation}
\label{app:reward_calculation}
For each quantitative record $\left([l_{ij},u_{ij}],a_{ij}\right)$, the absolute deviation $d_{ij}$ of the measured value $a_{ij}$ from the permissible bound $[l_{ij}, u_{ij}]$ is calculated as:
\begin{equation}
d_{ij}
=
\max\left\{
l_{ij}-a_{ij},\,
a_{ij}-u_{ij},\,
0
\right\}
\end{equation}
Thus, $d_{ij}=0$ when $a_{ij}\in[l_{ij},u_{ij}]$; otherwise, it is the distance to the nearest range boundary.
We first define $\bar{l}_{ij}=\max(l_{ij},0)$ and $\bar{u}_{ij}=\max(u_{ij},0)$, and calculate the characteristic range magnitude $s_{ij}$ of $[l_{ij}, u_{ij}]$ as follows:
\begin{equation}
s_{ij}
=
\begin{cases}
\max\left(1,\frac{\bar l_{ij}+\bar u_{ij}}{2}\right),
& u_{ij}<+\infty\\[3pt]
\max\left(1,\bar l_{ij}\right),
& u_{ij}=+\infty
\end{cases}
\end{equation}
In addition, we employ a binary format reward that equals zero when a response is truncated at the maximum generation length or contains an unclosed \texttt{<think>} block, and one otherwise. 
When a reasoning block is present, only the final-answer text is used for constraint verification. 
The final training reward is obtained by multiplying this format reward by the hierarchical reward defined in Section~\ref{sec:graded_constraint_reward_modeling}:
\begin{equation}
R_{\mathrm{final}}(x,y)
=
R_{\mathrm{format}}(y)\cdot R(x,y)
\end{equation}
For a fair comparison, we apply the same format reward to the RL-ILA and RL-CLA baselines.

\section{Details of Experiments}
\subsection{Evaluation Metrics}
\label{app:evaluation_metrics}
We evaluate precise instruction-following across four benchmarks: our test set, IFEval \citep{zhou2023instruction}, IFBench \citep{pyatkin2026generalizing}, and IFHierBench \citep{mao2026ifhierbench}. 
IFEval and IFBench employ four metrics: prompt- and instruction-level accuracy under both strict and loose evaluation conditions (P/I-S/L). 
Prompt-level accuracy requires all constraints to be satisfied, whereas instruction-level accuracy measures the proportion of satisfied constraints. 
Loose evaluation applies predefined normalizations to the response before verification. 
IFHierBench utilizes only prompt- and instruction-level strict accuracy (P/I-S). 
For our test set, following ImpRIF \citep{yang-etal-2026-imprif}, we report CSR (Constraint Success Rate) and ISR (Instruction Success Rate). 
Given $N$ instances, where the $i$-th instance contains an instruction $x_i$ with constraints $C_i$ and a corresponding response $y_i$, these metrics are computed as:
\begin{equation}
\mathrm{CSR}
=
\frac{1}{N}\sum_{i=1}^{N}
\frac{1}{|C_i|}
\sum_{c\in C_i} I(x_i,y_i,c)
\label{eq:csr}
\end{equation}
\begin{equation}
\mathrm{ISR}
=
\frac{1}{N}\sum_{i=1}^{N}
\prod_{c\in C_i} I(x_i,y_i,c)
\label{eq:isr}
\end{equation}
where $I(x_i,y_i,c)$ is an indicator function denoting whether response $y_i$ satisfies constraint $c$. 
Finally, for general benchmarks MATH-500 \citep{lightman2024let}, GPQA \citep{rein2023gpqa}, and MMLU-Pro \citep{wang2024mmlu}, we report standard answer accuracy.

\subsection{Implementation Details}
\label{app:training_details}
For DPO training, we use the Llama-Factory \citep{zheng2024llamafactory} framework, sampling 5 responses per instruction with a temperature of 1.0 and a top-p value of 0.9. 
The responses with the highest and lowest rewards are used to construct preference pairs. 
We use the Zero Redundancy Optimizer (ZeRO) \citep{rajbhandari2020zero} stage 3 from the DeepSpeed library \citep{rasley2020deepspeed} and the AdamW \citep{kingma2014adam, loshchilov2017decoupled} optimizer with a weight decay of 0.1.
The maximum sequence length is set to 8192 tokens.
The peak learning rate is set to 1e-6 with a 10\% warmup ratio and a cosine scheduler. 
The batch size is 128.
Training utilizes a sigmoid loss function with a beta value of 0.1 and spans 1 epoch.
Following previous works \citep{hou2024chatglm}, an additional SFT loss is added to the chosen response with a weight of 0.1.
The reward calculation and DPO training are both conducted on 8 H800 GPUs.

For GRPO training, we set the learning rate to 2e-6, with the maximum sequence length 4096 for prompts and 8192 for responses. 
The train-batch size and PPO mini-batch size are each set to 32. 
For rollout response generation, we employ the vllm \citep{kwon2023efficient} framework, utilizing 50\% of the available GPU memory,  with a temperature and top-p value of 1.0.
To encourage stable training, we also incorporate KL divergence regularization with a coefficient of 0.001, using the low-variance KL implementation. 
We save checkpoints every 30 steps during training. 
Following TULU 3 \citep{lambert2024t} and VerIF \citep{peng-etal-2025-verif}, we use IFEval \citep{zhou2023instruction} as the validation set to select the best checkpoint.
All experiments are conducted on 32 H800 GPUs, with 8 GPUs dedicated to model training and the remaining 24 GPUs allocated for API deployment of the judge model.
The training process runs for one epoch, taking approximately 50 hours.

\subsection{Detailed Results of Each Constraint Category}
\label{app:detailed_results}
\input{tables/results_per_scope}
\input{tables/results_per_target}
\input{tables/results_per_range}
Tables \ref{tab:results_per_scope}, \ref{tab:results_per_target} and \ref{tab:results_per_range} present the CSR of each \textit{Scope}, \textit{Target} and \textit{Range} category on our test set, respectively.
For \textit{Scope}, LLMs generally perform best on Global constraints, while their performance drops markedly on single-layer scope selectors (i.e., Traversal, Position, and Condition), and reaches its lowest point on recursively composed Nested scopes, demonstrating that scope complexity poses a growing challenge to LLMs.
For \textit{Target}, Capacity and Relations consistently emerge as the most challenging categories, suggesting that LLMs still struggle with precise control over text length and unit counts, as well as coordinating dependencies among multiple objects.
For \textit{Range}, One-Sided Bounds and Prohibition are comparatively easier, whereas Equality and Intervals pose greater challenges because they demand more precise numerical control. 
Notably, Arithmetic Relations prove particularly difficult, as they require models to maintain exact quantitative relationships among multiple targets, exposing a persistent weakness in jointly tracking and controlling coupled quantities across different response segments.

\subsection{Details of Constraint Verification Methods}
\label{app:constraint_verification}
For pure LLM judges, we modify the judge prompt in Table \ref{tab:tool_grounded_judgement} to exclude auxiliary verification results, requiring 
the judge model to directly output the final judgment alongside the detailed statistical information for each counting object.
In instances where there is a discrepancy between the judgment result and the statistical information, we adopt the statistical information as the definitive result.
For the rule-based verifier, we adjust the tool design prompt in Table \ref{tab:tool_design}, requiring the generated tool to take the entire response as input and yield a binary judgment on constraint satisfaction. To calculate pairwise agreement, we assign \textit{win}, \textit{tie}, or \textit{lose} labels to a pair of responses for a given instruction based on the number of satisfied constraints evaluated by either humans or the judge model, and compute their agreement accordingly.

\section{Limitations}
\input{sections/limitation}

\clearpage
\input{tables/atomic_constraint_generation_prompt}
\vspace{2em}
\input{tables/constraint_crossover_prompt}
\vspace{2em}
\input{tables/instruction_quality_assessment_prompt}
\vspace{2em}
\input{tables/constraint_checklist_quality_assessment_prompt}
\vspace{2em}
\input{tables/target_extraction_prompt}
\vspace{2em}
\input{tables/target_quality_assessment_prompt}
\vspace{2em}
\input{tables/tool_design_prompt}
\vspace{2em}
\input{tables/judge_prompt}

%% file: tables/results_per_scope.tex
\begin{table*}[!t]
% \centering % Replaced by the official center environment below.
\small
\caption{CSR (\%) across five \textit{Scope} categories on our test set. The highest result for each backbone model is \textbf{bolded}, and the overall highest is \underline{underlined}.}
\label{tab:results_per_scope}
\begin{center}
\begin{tabular}{l|ccccc}
\toprule
\textbf{Model} & \textbf{Global} & \textbf{Traversal} & \textbf{Position} & \textbf{Condition} & \textbf{Nested} \\
\midrule
\multicolumn{6}{c}{\textit{Frontier Models}} \\
\midrule
Seed-2.0-Pro
& \underline{75.1}
& \underline{59.8}
& \underline{70.8}
& \underline{63.5}
& \underline{64.9} \\
Gemini-2.5-Pro
& 71.5 & 56.6 & 60.8 & 61.0 & 56.9 \\
DeepSeek-V3.2
& 61.5 & 45.7 & 45.1 & 44.2 & 43.1 \\
GPT-4.1
& 62.5 & 49.2 & 49.0 & 46.3 & 45.6 \\
QwQ-32B
& 58.3 & 40.4 & 45.5 & 43.3 & 39.9 \\
\midrule
\multicolumn{6}{c}{\textit{Our Models}} \\
\midrule
Qwen3-4B
& 52.0 & 35.1 & 35.9 & 36.2 & 31.3 \\
\quad + DPO
& 53.4 & 39.9 & 39.8 & 38.1 & 34.5 \\
\quad + RL-ILA
& 56.9 & 45.9 & 43.4 & 46.2 & 42.9 \\
\quad + RL-CLA
& 63.1 & 48.3 & 48.9 & 49.8 & 48.7 \\
\rowcolor{rowblue} \quad + ScopeIF (Ours)
& \textbf{65.6}
& \textbf{53.8}
& \textbf{53.2}
& \textbf{59.0}
& \textbf{56.3} \\
\midrule
Qwen3-8B
& 53.7 & 40.8 & 39.5 & 39.0 & 32.7 \\
\quad + DPO
& 54.5 & 41.5 & 40.3 & 41.3 & 35.9 \\
\quad + RL-ILA
& 61.3 & 48.0 & 50.5 & 51.3 & 49.1 \\
\quad + RL-CLA
& 66.4 & 56.1 & 57.2 & 56.0 & 56.0 \\
\rowcolor{rowblue} \quad + ScopeIF (Ours)
& \textbf{67.2}
& \textbf{57.1}
& \textbf{61.0}
& \textbf{60.1}
& \textbf{61.2} \\
\bottomrule
\end{tabular}
\end{center}
% Original bottom caption retained; the active caption is above the table.
% \caption{CSR (\%) across five \textit{Scope} categories on our test set. The highest result for each backbone model is \textbf{bolded}, and the overall highest is \underline{underlined}.}
% \label{tab:results_per_scope}
\end{table*}

%% file: tables/results_per_target.tex
\begin{table*}[!t]
% \centering % Replaced by the official center environment below.
\small
\caption{CSR (\%) across six \textit{Target} categories on our test set. }
\label{tab:results_per_target}
\begin{center}
\begin{tabular}{l|cccccc}
\toprule
\textbf{Model} & \textbf{Capacity} & \textbf{Literals} & \textbf{Structure} & \textbf{Checklist} & \textbf{Pattern} & \textbf{Relations} \\
\midrule
\multicolumn{7}{c}{\textit{Frontier Models}} \\
\midrule
Seed-2.0-Pro & \underline{61.4} & \underline{69.5} & \underline{79.1} & \underline{74.2} & \underline{74.6} & \underline{58.2} \\
Gemini-2.5-Pro & 55.5 & 64.6 & 73.8 & 68.3 & 69.8 & 51.5 \\
DeepSeek-V3.2 & 43.8 & 52.2 & 63.4 & 52.6 & 57.2 & 35.2 \\
GPT-4.1 & 45.6 & 51.2 & 69.6 & 55.7 & 61.8 & 37.3 \\
QwQ-32B & 39.1 & 47.6 & 64.3 & 52.7 & 53.7 & 36.6 \\
\midrule
\multicolumn{7}{c}{\textit{Our Models}} \\
\midrule
Qwen3-4B & 35.4 & 41.9 & 47.1 & 44.0 & 46.9 & 26.8 \\
\quad + DPO & 37.6 & 44.2 & 53.7 & 49.0 & 47.6 & 28.6 \\
\quad + RL-ILA & 42.3 & 49.3 & 58.1 & 55.5 & 53.4 & 38.1 \\
\quad + RL-CLA & 44.2 & 55.8 & 68.8 & 58.5 & 62.8 & 39.7 \\
\rowcolor{rowblue} \quad + ScopeIF (Ours) & \textbf{47.8} & \textbf{59.9} & \textbf{73.1} & \textbf{67.3} & \textbf{67.3} & \textbf{50.2} \\
\midrule
Qwen3-8B & 36.5 & 44.6 & 52.5 & 47.6 & 48.5 & 29.7 \\
\quad + DPO & 38.3 & 44.5 & 57.8 & 48.4 & 51.5 & 30.1 \\
\quad + RL-ILA & 44.4 & 54.4 & 65.3 & 60.5 & 61.8 & 43.6 \\
\quad + RL-CLA & 49.4 & 60.9 & 72.1 & 67.8 & 69.7 & 47.1 \\
\rowcolor{rowblue} \quad + ScopeIF (Ours) & \textbf{53.7} & \textbf{65.1} & \textbf{72.9} & \textbf{69.4} & \textbf{70.8} & \textbf{50.4} \\
\bottomrule
\end{tabular}
\end{center}
% Original bottom caption retained; the active caption is above the table.
% \caption{CSR (\%) across six \textit{Target} categories on our test set. }
% \label{tab:results_per_target}
\end{table*}

%% file: tables/results_per_range.tex
\begin{table*}[!t]
% \centering % Replaced by the official center environment below.
\small
\caption{CSR (\%) across six \textit{Range} categories on our test set. }
\label{tab:results_per_range}
\begin{center}
% Match the visible caption-to-rule gap in Tables 6 and 7.
\vspace{-2.75mm}
\resizebox{\textwidth}{!}{
\begin{tabular}{l|cccc|cc}
\toprule
\multirow{2}{*}{\textbf{Model}} & \multicolumn{4}{c|}{\textbf{Absolute Range}} & \multicolumn{2}{c}{\textbf{Relative Range}} \\
\cmidrule{2-7}
& \textbf{One-Sided Bounds} & \textbf{Prohibition} & \textbf{Equality} & \textbf{Intervals} & \textbf{Comparison} & \textbf{Arithmetic Relations} \\
\midrule
\multicolumn{7}{c}{\textit{Frontier Models}} \\
\midrule
Seed-2.0-Pro & \underline{80.3} & \underline{82.0} & \underline{61.0} & \underline{61.5} & 68.9 & \underline{40.2} \\
Gemini-2.5-Pro & 76.8 & 80.9 & 50.2 & 56.1 & \underline{70.6} & 32.2 \\
DeepSeek-V3.2 & 68.1 & 74.2 & 34.5 & 41.8 & 47.9 & 15.0 \\
GPT-4.1 & 70.8 & 76.3 & 37.0 & 41.2 & 57.0 & 16.3 \\
QwQ-32B & 62.4 & 70.4 & 33.8 & 37.9 & 49.0 & 22.7 \\
\midrule
\multicolumn{7}{c}{\textit{Our Models}} \\
\midrule
Qwen3-4B & 56.7 & 65.4 & 26.8 & 27.9 & 38.7 & 12.3 \\
\quad + DPO & 60.4 & 68.8 & 30.1 & 28.4 & 42.2 & 12.3 \\
\quad + RL-ILA & 63.9 & 72.1 & 34.8 & 38.0 & 51.0 & 22.7 \\
\quad + RL-CLA & 67.6 & 72.8 & 41.8 & 44.8 & 56.1 & 29.1 \\
\rowcolor{rowblue} \quad + ScopeIF (Ours) & \textbf{72.9} & \textbf{78.0} & \textbf{47.2} & \textbf{50.6} & \textbf{65.5} & \textbf{30.7} \\
\midrule
Qwen3-8B & 61.0 & 66.9 & 28.2 & 30.0 & 43.0 & 12.0 \\
\quad + DPO & 61.5 & 68.3 & 30.4 & 30.9 & 47.0 & 16.9 \\
\quad + RL-ILA & 67.1 & 75.3 & 41.3 & 45.2 & 53.6 & 30.4 \\
\quad + RL-CLA & 72.6 & 76.1 & 49.2 & 51.1 & 62.7 & \textbf{37.4} \\
\rowcolor{rowblue} \quad + ScopeIF (Ours) & \textbf{75.2} & \textbf{79.2} & \textbf{52.8} & \textbf{56.5} & \textbf{63.8} & 33.4 \\
\bottomrule
\end{tabular}
}
\end{center}
% Original bottom caption retained; the active caption is above the table.
% \caption{CSR (\%) across six \textit{Range} categories on our test set. }
% \label{tab:results_per_range}
\end{table*}

%% file: sections/limitation.tex
The limitations of our work are summarized as follows:

\paragraph{Potential Bias in Synthetic Data Construction.}
The construction of \textsc{ScopeInstruct} primarily relies on Seed-2.0-Pro for constraint generation and instruction synthesis, which may introduce model-specific preferences.
Although we employ a different LLM, GPT-OSS-120B, for quality inspection to mitigate this issue, incorporating multiple data-generation models and broader human auditing may further improve data robustness and diversity.
We consider this an important direction for future work.

\paragraph{Cost of Tool-Grounded Verification.}
ScopeIF employs judge models to dynamically design response-specific tools for constraint verification, introducing additional computational overhead during reward calculation.
We consider this overhead a necessary trade-off for reliably isolating relevant response segments and measuring complex targets that elude both a fixed inventory of rule-based verifiers and pure LLM judges.
Importantly, this additional cost is incurred only during the training stage and does not affect the inference efficiency of the resulting policy models.
Nonetheless, further reducing this overhead through lightweight judge models, batched verification, and reusable verification tools for recurring constraint patterns remains an important future work.

%% file: tables/atomic_constraint_generation_prompt.tex
{\scriptsize
\setlength{\parindent}{0pt}%
\setlength{\parskip}{0pt}%
\captionof{table}{The prompt template for atomic constraint generation.}
\label{tab:atomic_constraint_generation}
\par\nopagebreak[4]\vspace{\baselineskip}
% \par\noindent\rule{\linewidth}{0.8pt}\par\vspace{2pt} % Original opening rule retained.
\par\noindent\rule{\linewidth}{0.8pt}\par\vspace{2pt}\nopagebreak[4]

I need to construct diverse and challenging objective constraints.
In this task, each objective constraint is represented by three dimensions: \textit{Scope}, \textit{Target}, and \textit{Range}.
They dictate the permissible bounds of specific measurable objects within designated response segments.
The taxonomy is provided as follows:

\textcolor{red}{\{constraint schema\}}

\vspace{\baselineskip}
I will provide you with a user instruction and a specific constraint category that contains the combination of these three dimensions.
Your task is to add one or more suitable objective constraints to the user instruction that conform to the specified constraint category to construct new instructions.
You must provide five distinct construction results.
I will also provide previously generated instructions for the same category.
You must ensure that the newly added constraints in the five results differ as much as possible from those in the previous results and remain diverse.
The specific constraint category will be provided in the following format:

\texttt{\textasciigrave\textasciigrave\textasciigrave}

\texttt{\{}

\hspace*{1em}\texttt{"Scope": ... (one of "Global", "Traversal", "Position", and "Condition"),}

\hspace*{1em}\texttt{"Nested Scope": ... (true or false),}

\hspace*{1em}\texttt{"Target": ... (one of "Capacity", "Literals", "Structure", "Checklist", "Pattern", and "Relations"),}

\hspace*{1em}\texttt{"Range": ... (one of "One-Sided Bounds", "Prohibition", "Equality", "Intervals", "Comparison", and "Arithmetic Relations")}

\texttt{\}}

\texttt{\textasciigrave\textasciigrave\textasciigrave}

When completing this task, you must follow the principles below:

1. To integrate the original instruction and new constraints more naturally and coherently, you can appropriately revise the instruction.
Objective constraints that are already present in the original instruction should preferably be removed or replaced.
Moreover, you should focus on adding objective constraints and avoid subjective constraints such as topic or style.
The newly constructed instruction must use the same language as the original instruction.

2. The newly added constraints must be natural, clear, understandable, reasonable, practical, and suitable for the new user instruction.
They must not be contrived or impossible to fulfill.
The objective constraints must not conflict with one another or with any other part of the instruction.

3. The new user instruction must contain complete information and must not omit any necessary content.
You must ensure that a model can understand the meaning and complete information of the new instruction and fulfill its task when the original instruction is not visible.

4. The newly added constraints must be sufficiently challenging for LLMs.
Thus, you should carefully design the \textit{Range} according to the characteristics of the user instruction.
As a general criterion, without the added constraint, LLMs' output should have a high probability of violating it, and satisfying the added constraint should require LLMs to adjust their response deliberately.

5. The newly added constraints in all five results must conform to the given constraint category.

6. The newly added constraints in all five results must differ from those in the previously generated instruction while also remaining diverse from each other.
The response unit, implementations of \textit{Scope} selector, secondary \textit{Target} categories, \textit{Range} values, and other unspecified details should be as diverse as possible.
The final constraints should also be expressed in diverse ways.
I will provide the distributions of response units and secondary \textit{Target} categories for constraints from previously constructed instructions.
You should prioritize constraints that have received less coverage when constructing the constraints for the current round.

7. The newly added constraints should use natural and varied expressions rather than being mechanically phrased as count-based statements.
A qualitative constraint (e.g., valid JSON format or a specific text ending) is a special case of an Equality \textit{Range} of one.

8. The newly added constraints should be concise and clear.
Avoid appending commonsense explanations, definitions, or examples.

9. You are not limited to adding only one constraint and are encouraged to add multiple constraints from the given category to construct a more complex constraint.
If multiple constraints are added, they must use closely related implementations either as parallel instances of the same atomic constraint or as constraints with a clear logical progression.
They must not be unrelated constraints with substantially different forms.
Their category must be the same.

\vspace{\baselineskip}
You must construct five high-quality new instructions and provide both the new instructions and the newly added constraints.
Strictly follow the output format below:

\vspace{\baselineskip}
\texttt{\textasciigrave\textasciigrave\textasciigrave python}

\textbf{[The Start of Analysis]}

... (Provide a detailed analysis and plan explaining how to construct new instructions by adding diverse, high-quality, challenging objective constraints that differ from previously generated constraints and conform to the specified constraint category)

\textbf{[The End of Analysis]}

\vspace{\baselineskip}
\textbf{[The Start of Construction Results]}

\texttt{[}

\hspace*{1.5em}\texttt{\{}

\hspace*{3em}\texttt{"Original Instruction": ... (string; directly copy the original instruction without any modifications),}

\hspace*{3em}\texttt{"New Instruction": ... (string; the constructed new instruction),}

\hspace*{3em}\texttt{"Newly Added Constraints": ... (string; all constraints added to the original instruction. Directly extract segments from the new instruction without any modifications. The extraction must be complete and cover all added constraints),}

\hspace*{3em}\texttt{"Scope": ... (string; use the format "Scope Selector(Response Unit)". The Scope Selector must be one of "Global", "Traversal", "Position", and "Condition". The Response Unit must be one of "Entire Response", "Structural Block", "Structural Element", "Paragraph", "Sentence", "Word", and "Character". Connect nested Scope with " -> "),} 

\hspace*{3em}\texttt{"Secondary Target Category": ... (string; one of Secondary Target Category)}

\hspace*{1em}\texttt{\},}

\hspace*{1em}\texttt{... (a list of dictionaries containing exactly five items)}

\texttt{]}

\textbf{[The End of Construction Results]}

\texttt{\textasciigrave\textasciigrave\textasciigrave}

\vspace{\baselineskip}
The user instruction, the specific constraint category, and the previously constructed instruction are provided below:

\textbf{[The Start of User Instruction]}

\textcolor{red}{\{instruction\}}

\textbf{[The End of User Instruction]}

\vspace{\baselineskip}
\textbf{[The Start of Constraint Category]}

\textcolor{red}{\{category\}}

\textbf{[The End of Constraint Category]}

\vspace{\baselineskip}
\textbf{[The Start of Previous Constructed Instructions]}

\textcolor{red}{\{existing\_instructions\}}

\textbf{[The End of Previous Constructed Instructions]}

\vspace{\baselineskip}
\textbf{[The Start of Previous Response Unit Distribution]}

\textcolor{red}{\{existing\_response\_unit\_distribution\}}

\textbf{[The End of Previous Response Unit Distribution]}

\vspace{\baselineskip}
\textbf{[The Start of Previous Secondary Target Category Distribution]}

\textcolor{red}{\{existing\_secondary\_target\_category\_distribution\}}

\textbf{[The End of Previous Secondary Target Category Distribution]}

% \par\vspace{2pt}\noindent\rule{\linewidth}{0.8pt}\par % Original closing rule retained.
\par\nopagebreak[4]\vspace{2pt}\noindent\rule{\linewidth}{0.8pt}\par
% Original bottom caption retained; the active caption is above the table.
% \captionof{table}{The prompt template for atomic constraint generation.}
% \label{tab:atomic_constraint_generation}
}

%% file: tables/constraint_crossover_prompt.tex
{\scriptsize
\setlength{\parindent}{0pt}%
\setlength{\parskip}{0pt}%
\captionof{table}{The prompt template for constraint crossover.}
\label{tab:constraint_crossover}
\par\nopagebreak[4]\vspace{\baselineskip}
% \par\noindent\rule{\linewidth}{0.8pt}\par\vspace{2pt} % Original opening rule retained.
\par\noindent\rule{\linewidth}{0.8pt}\par\vspace{2pt}\nopagebreak[4]

I need to construct diverse and challenging objective constraints.
In this task, each objective constraint is represented by three dimensions: \textit{Scope}, \textit{Target}, and \textit{Range}.
They dictate the permissible bounds of specific measurable objects within designated response segments.
The taxonomy is provided as follows:

\textcolor{red}{\{constraint schema\}}

\vspace{\baselineskip}
I will provide you with a user instruction and several objective constraints.
Your task is to construct five more complex new instructions by adding suitable objective constraints to the user instruction in distinct ways.
For each new instruction, you must add at least \textcolor{red}{\{constraint\_number\}} constraints.
The same constraint may be appropriately modified and added multiple times.
The constructed instructions should be diverse and differ as much as possible. 
Each given constraint consists of an ID, its content, and its classification under the taxonomy described above.
The format is as follows:

\texttt{\textasciigrave\textasciigrave\textasciigrave}

\texttt{\{}

\hspace*{1.5em}\texttt{"Constraint ID": ... (integer),}

\hspace*{1.5em}\texttt{"Constraint Content": ... (string),}

\hspace*{1.5em}\texttt{"Constraint Classification": \{}

\hspace*{3em}\texttt{"Scope": ... (string),}

\hspace*{3em}\texttt{"Primary Target": ... (one of "Capacity", "Literals", "Structure", "Checklist", "Pattern", and "Relations"),}

\hspace*{3em}\texttt{"Secondary Target": ... (string),}

\hspace*{3em}\texttt{"Range": ... (one of "One-Sided Bounds", "Prohibition", "Equality", "Intervals", "Comparison", and "Arithmetic Relations")}

\hspace*{1.5em}\texttt{\}}

\texttt{\}}

\texttt{\textasciigrave\textasciigrave\textasciigrave}

When completing this task, you must follow the principles below:

1. To integrate the original instruction and new constraints more naturally and coherently, you can appropriately revise the instruction.
Objective constraints that are already present in the original instruction should preferably be removed or replaced.
Every newly added constraint must be obtained by adjusting or modifying one of the given constraints.
You are prohibited from inventing new constraints unrelated to the given constraints.
Moreover, you should focus on adding objective constraints and avoid subjective constraints such as topic or style.

2. The newly constructed instructions must use the same language as the original user instruction.
You must not add constraints that are specific to other languages and different from the newly constructed instructions.

3. The new user instruction must contain complete information and must not omit any necessary content.
You must ensure that a model can understand the meaning and complete information of the new instruction and fulfill its task when the original instruction is not visible.

4. The newly added constraints must be natural, clear, understandable, reasonable, practical, and suitable for the new user instruction.
They must not be contrived or impossible to fulfill.
The objective constraints must not conflict with one another or with any other part of the instruction.

5. The newly added constraints must be sufficiently challenging for LLMs.
Thus, you should carefully design the \textit{Range} according to the characteristics of the user instruction.
As a general criterion, without the added constraint, LLMs' output should have a high probability of violating it, and satisfying the added constraint should require LLMs to adjust their response deliberately.

6. The newly added constraints in all five results must be as diverse as possible and should collectively cover a diverse selection of the given constraints.
The final constraints should also be expressed in diverse ways.

7. The newly added constraints should be concise and clear.
Avoid appending commonsense explanations, definitions, or examples.

8. To reiterate, the same given constraint is not limited to being added only once. You are encouraged to modify and add the same constraint multiple times to construct more complex constraint forms.

\vspace{\baselineskip}
You must generate five high-quality new user instructions and provide each new instruction together with all objective constraints it contains.
Strictly follow the output format below:

\vspace{\baselineskip}
\texttt{\textasciigrave\textasciigrave\textasciigrave}

\textbf{[The Start of Analysis]}

... (Provide a detailed analysis and plan explaining how to construct new instructions by adding diverse, high-quality, and challenging new constraints)

\textbf{[The End of Analysis]}

\vspace{\baselineskip}
\textbf{[The Start of Construction Results]}

\texttt{[}

\hspace*{1.5em}\texttt{\{}

\hspace*{3em}\texttt{"Instruction ID": ... (integer from 1 to 5),}

\hspace*{3em}\texttt{"New User Instruction": ... (a new user instruction constructed by adding suitable constraints to the original instruction),}

\hspace*{3em}\texttt{"All Added Constraints in the New User Instruction": [}

\hspace*{4.5em}\texttt{\{}

\hspace*{6em}\texttt{"Constraint Content": ... (string; completely extract one newly added constraint contained in the new instruction),}

\hspace*{6em}\texttt{"Source Constraint ID": ... (integer)}

\hspace*{4.5em}\texttt{\},}

\hspace*{4.5em}\texttt{... (extract all newly added constraints in the new user instruction without any omissions. If the same given constraint is used multiple times, output multiple items with the same Source Constraint ID rather than merging them into one item)}

\hspace*{3em}\texttt{]}

\hspace*{1.5em}\texttt{\},}

\hspace*{1.5em}\texttt{... (a list containing exactly five items, each of which contains a new user instruction and its newly added constraints)}

\texttt{]}

\textbf{[The End of Construction Results]}

\texttt{\textasciigrave\textasciigrave\textasciigrave}

\vspace{\baselineskip}
The user instruction and given constraints are provided below:

\textbf{[The Start of User Instruction]}

\textcolor{red}{\{instruction\}}

\textbf{[The End of User Instruction]}

\vspace{\baselineskip}
\textbf{[The Start of Given Constraints]}

\textcolor{red}{\{given\_constraints\}}

\textbf{[The End of Given Constraints]}

% \par\vspace{2pt}\noindent\rule{\linewidth}{0.8pt}\par % Original closing rule retained.
\par\nopagebreak[4]\vspace{2pt}\noindent\rule{\linewidth}{0.8pt}\par
% Original bottom caption retained; the active caption is above the table.
% \captionof{table}{The prompt template for constraint crossover.}
% \label{tab:constraint_crossover}
}

%% file: tables/instruction_quality_assessment_prompt.tex
{\scriptsize
\setlength{\parindent}{0pt}%
\setlength{\parskip}{0pt}%
\captionof{table}{The prompt template for user instruction quality assessment.}
\label{tab:instruction_quality_assessment}
\par\nopagebreak[4]\vspace{\baselineskip}
% \par\noindent\rule{\linewidth}{0.8pt}\par\vspace{2pt} % Original opening rule retained.
\par\noindent\rule{\linewidth}{0.8pt}\par\vspace{2pt}\nopagebreak[4]

You are a professional artificial intelligence expert proficient in machine learning and natural language processing, with expertise in assessing the quality of user instructions.
I will provide you with a user instruction. 
You must carefully analyze its logic and assess its quality.
When assessing the quality of a user instruction, a high-quality instruction must satisfy all of the following conditions. 
Violation of any one condition means that the instruction should be considered low-quality.
You must carefully analyze the instruction word by word to determine whether it satisfies each of the following conditions:

\vspace{\baselineskip}
1. The intent is completely explicit.

2. The semantics are completely clear, and the language is coherent, fluent, and easy to understand, without ambiguity, incomprehensible statements, or grammatical errors.

3. It contains no erroneous content or incorrect information.

4. It does not exceed the capabilities of an offline text-only model; an offline text-only model must be able to complete the task. 
Being offline means that the model cannot access niche knowledge, such as obscure books, files, materials, or other information, nor can it access specialized knowledge, such as professional engineering knowledge or specialized books.

5. It contains no impossible constraints. Note that unconventional, highly difficult constraints are permitted, provided that they are not impossible to fulfill.

6. It contains no erroneous or disorganized logic.

7. Its internal logic is consistent, with no contradictory logic or constraints. No parts of the instruction conflict with or contradict one another, and all constraints can be fulfilled simultaneously.

8. The instruction and its accompanying materials are complete, with no missing content or data.

9. The user instruction provides all necessary data and information.

10. The user instruction must not contain any problem that requires multi-turn interaction to resolve. In other words, the AI assistant must be able to respond to the instruction completely within a single turn.

11. The user instruction is concise without redundant explanations.

\vspace{\baselineskip}
You are prohibited from answering the user instruction. You only need to determine whether it is high-quality or low-quality.
Strictly follow the format below when providing your analysis and conclusion:

\vspace{\baselineskip}
\texttt{\textasciigrave\textasciigrave\textasciigrave}

\textbf{Analysis:} ... (Provide a comprehensive, detailed, and meticulous analysis)

\textbf{Final Conclusion:} ... (The final conclusion must be either ``[[High Quality]]'' or ``[[Low Quality]]'')

\texttt{\textasciigrave\textasciigrave\textasciigrave}

\vspace{\baselineskip}
\noindent\begin{minipage}{\linewidth}
Here is the user instruction:

\textbf{[The Start of User Instruction]}

\textcolor{red}{\{instruction\}}

\textbf{[The End of User Instruction]}

% \par\vspace{2pt}\noindent\rule{\linewidth}{0.8pt}\par % Original closing rule retained.
\par\nopagebreak[4]\vspace{2pt}\noindent\rule{\linewidth}{0.8pt}\par
% Original bottom caption retained; the active caption is above the table.
% \captionof{table}{The prompt template for user instruction quality assessment.}
% \label{tab:instruction_quality_assessment}
\end{minipage}
}

%% file: tables/constraint_checklist_quality_assessment_prompt.tex
{\scriptsize
\setlength{\parindent}{0pt}%
\setlength{\parskip}{0pt}%
\captionof{table}{The prompt template for constraint checklist quality assessment.}
\label{tab:constraint_checklist_quality_assessment}
\par\nopagebreak[4]\vspace{\baselineskip}
\par\noindent\rule{\linewidth}{0.8pt}\par\nopagebreak[4]\vspace{2pt}

You are a professional artificial intelligence expert proficient in machine learning and natural language processing, with expertise in assessing the quality of user instructions.
I will provide you with a user instruction containing objective constraints, together with the extraction results for those constraints.
You must carefully analyze and thoroughly examine whether the extraction results are correct.
Objective constraints dictate the permissible bounds of specific measurable objects within designated response segments.
The taxonomy is provided as follows:
\textcolor{red}{\{constraint schema\}}

\vspace{\baselineskip}
A high-quality extraction result must satisfy all of the following conditions. 
Violation of any one condition means that the extraction result should be considered low-quality.
You must carefully analyze the extraction results word by word to determine whether they satisfy each of the following conditions:

\vspace{\baselineskip}
1. All objective constraints in the user instruction are completely extracted, without containing any other constraints, and none of them are omitted.

2. All objective constraints in the user instruction are accurately extracted, and the meaning of each extracted constraint is completely consistent with the instruction.

3. Each objective constraint is extracted exactly once, with no duplicate extractions.

4. Each extracted objective constraint is expressed clearly, without any ambiguity.

5. There is no restriction on extraction granularity. Multiple constraints may either be combined into a single extracted item or divided into multiple extracted items.

\vspace{\baselineskip}
You are prohibited from answering the user instruction. You only need to determine whether the extraction result is high-quality or low-quality.
Strictly follow the format below when providing your analysis and conclusion:

\vspace{\baselineskip}
\texttt{\textasciigrave\textasciigrave\textasciigrave}

\textbf{Analysis:} ... (Provide a comprehensive, detailed, and meticulous analysis)

\textbf{Final Conclusion:} ... (The final conclusion must be either ``[[High Quality]]'' or ``[[Low Quality]]'')

\texttt{\textasciigrave\textasciigrave\textasciigrave}

\vspace{\baselineskip}
The user instruction and its corresponding objective constraint extraction results are provided below:

\textbf{[The Start of User Instruction]}

\textcolor{red}{\{prompt\}}

\textbf{[The End of User Instruction]}

\vspace{\baselineskip}
\textbf{[The Start of Extraction Results]}

\textcolor{red}{\{constraint\_checklist\}}

\textbf{[The End of Extraction Results]}

% \par\vspace{2pt}\noindent\rule{\linewidth}{0.8pt}\par % Original closing rule retained.
\par\nopagebreak[4]\vspace{2pt}\noindent\rule{\linewidth}{0.8pt}\par
% Original bottom caption retained; the active caption is above the table.
% \captionof{table}{The prompt template for constraint checklist quality assessment.}
% \label{tab:constraint_checklist_quality_assessment}
}

%% file: tables/target_extraction_prompt.tex
{\scriptsize
\setlength{\parindent}{0pt}%
\setlength{\parskip}{0pt}%
\captionof{table}{The prompt template for target extraction for each constraint.}
\label{tab:target_extraction}
\par\nopagebreak[4]\vspace{\baselineskip}
% \par\noindent\rule{\linewidth}{0.8pt}\par\vspace{2pt} % Original opening rule retained.
\par\noindent\rule{\linewidth}{0.8pt}\par\vspace{2pt}\nopagebreak[4]

You are an expert skilled at extracting important information from instructions.
I will provide you with a user instruction and a given constraint contained in that instruction.
My goal is to determine whether some responses satisfy the given constraint.
Your task is to completely extract all counting objects of the given constraint.
You must follow the principles below when completing this task:

\vspace{\baselineskip}
1. Counting objects can take two forms.
Every counting object must be expressed using the first form whenever possible.
The second form may be used only when the counting object cannot be expressed using the first form.

\hspace*{1.5em}a. If the subconstraint directly dictates the quantity of certain content, the counting object should be the quantity of that content, excluding the specific quantitative constraint.

\hspace*{1.5em}b. If the subconstraint cannot be expressed using the above form, the counting object should be expressed in a qualitative form: a question that can be answered with ``yes'' or ``no'' and indicates whether the corresponding subconstraint is satisfied.

2. The counting object must be divided comprehensively and without omission, and every part of the given constraint that needs to be verified must have a corresponding counting object.
The counting object must correspond exactly to the content of the given constraint.
Every subconstraint must have a corresponding counting object, and the counting object must not contain anything unrelated to the given constraint.
In other words, every counting object must fall within the scope of the given constraint, and the counting objects must collectively constitute a necessary and sufficient condition for satisfying the given constraint.
Furthermore, the counting object should be divided as atomically as possible, and complex constraints should be divided into multiple counting objects whenever possible.

3. You must distinguish whether each counting object is universal or specific.
``Universal'' means that the corresponding subconstraint applies to an unspecified number of multiple objects, whereas ``specific'' means that it applies to exactly one particular object.
You must provide the type of every counting object in the final output.
A specific counting object must correspond to the statistical result of exactly one particular object and must not correspond to multiple statistical results.

4. If the quantity in a constraint is expressed using the quantity of another reference object rather than a specific number, use the subject of the constraint as the counting object.
The reference value must not be treated as a counting object.

5. The counting object must use the same language as the user instruction.

\vspace{\baselineskip}
Use the following output format:

\vspace{\baselineskip}
\texttt{\textasciigrave\textasciigrave\textasciigrave\textasciigrave}

\textbf{[The Start of the Constraint]}

... (It must be identical to the constraint I provide, without any modification)

\textbf{[The End of the Constraint]}

\vspace{\baselineskip}
\textbf{[The Start of Counting Object]}

\texttt{\textasciigrave\textasciigrave\textasciigrave json}

\texttt{[}

\hspace*{1em}\texttt{\{}

\hspace*{2em}\texttt{"Counting Object": ... (Describe the counting object in detail, using the original wording of the constraint wherever possible),}

\hspace*{2em}\texttt{"Type": ... ("Universal" or "Specific")}

\hspace*{1em}\texttt{\},}

\hspace*{1em}\texttt{...}

\texttt{]}

\texttt{\textasciigrave\textasciigrave\textasciigrave}

\textbf{[The End of Counting Object]}

\texttt{\textasciigrave\textasciigrave\textasciigrave\textasciigrave}

\vspace{\baselineskip}
The user instruction and the given constraint contained in it are provided below:

\textbf{[The Start of User Instruction]}

\textcolor{red}{\{instruction\}}

\textbf{[The End of User Instruction]}

\vspace{\baselineskip}
\textbf{[The Start of the Constraint]}

\textcolor{red}{\{constraint\}}

\textbf{[The End of the Constraint]}

% \par\vspace{2pt}\noindent\rule{\linewidth}{0.8pt}\par % Original closing rule retained.
\par\nopagebreak[4]\vspace{2pt}\noindent\rule{\linewidth}{0.8pt}\par
% Original bottom caption retained; the active caption is above the table.
% \captionof{table}{The prompt template for target extraction for each constraint.}
% \label{tab:target_extraction}
}

%% file: tables/target_quality_assessment_prompt.tex
{\scriptsize
\setlength{\parindent}{0pt}%
\setlength{\parskip}{0pt}%
\captionof{table}{The prompt template for target extraction quality assessment.}
\label{tab:target_extraction_quality_assessment}
\par\nopagebreak[4]\vspace{\baselineskip}
% \par\noindent\rule{\linewidth}{0.8pt}\par\vspace{2pt} % Original opening rule retained.
\par\noindent\rule{\linewidth}{0.8pt}\par\vspace{2pt}\nopagebreak[4]

You are a professional artificial intelligence expert proficient in machine learning and natural language processing, with expertise in assessing the quality of user instructions.
I will provide you with a user instruction, a given constraint contained in that instruction, and a list of counting objects for this constraint.
Your task is to carefully analyze whether the object list is completely correct and satisfies all of the following conditions, and ultimately conclude with either ``[[High Quality]]'' or ``[[Low Quality]]''.
High-quality counting object lists must satisfy all of the following conditions.
Violation of any one condition means that the list should be considered low-quality.
You must carefully analyze the object list word by word to determine whether it satisfies each of the following
conditions:

\vspace{\baselineskip}
\makebox[1.5em][l]{1.}The statistical result of every counting object must be representable either by a number or by ``yes/no''.

\makebox[1.5em][l]{2.}Every counting object must be correctly classified as either ``Universal'' or ``Specific''.
``Universal'' means that the corresponding subconstraint applies to an unspecified number of multiple objects, whereas ``Specific'' means that it applies to exactly one particular object.
A specific counting object must not correspond to multiple statistical results.
The type of every counting object must be provided.

\makebox[1.5em][l]{3.}If the quantity in a constraint is expressed using the quantity of another reference object rather than a specific number, use the subject of the constraint as the counting object.
The reference value must not be treated as a counting object.

\makebox[1.5em][l]{4.}The counting objects must correspond exactly to the content of the given constraint.
Every subconstraint of the given constraint must have a corresponding counting object, and the counting objects must not contain anything unrelated to that constraint.
In other words, every counting object must fall within the scope of the given constraint, and the counting objects must collectively constitute a necessary and sufficient condition for satisfying the given constraint.

\vspace{\baselineskip}
Strictly follow the format below when providing your analysis and conclusion:

\vspace{\baselineskip}
\texttt{\textasciigrave\textasciigrave\textasciigrave}

\textbf{Analysis:} ... (Provide a comprehensive, detailed, and meticulous analysis)

\textbf{Final Conclusion:} ... (The final conclusion must be either ``[[High Quality]]'' or ``[[Low Quality]]'').

\texttt{\textasciigrave\textasciigrave\textasciigrave}

\vspace{\baselineskip}
The user instruction, the corresponding constraint, and the counting object list are provided below:

\textbf{[The Start of User Instruction]}

\textcolor{red}{\{instruction\}}

\textbf{[The End of User Instruction]}

\vspace{\baselineskip}
\textbf{[The Start of the Constraint]}

\textcolor{red}{\{constraint\}}

\textbf{[The End of the Constraint]}

\vspace{\baselineskip}
\textbf{[The Start of Counting Object List]}

\textcolor{red}{\{target\}}

\textbf{[The End of Counting Object List]}

% \par\vspace{2pt}\noindent\rule{\linewidth}{0.8pt}\par % Original closing rule retained.
\par\nopagebreak[4]\vspace{2pt}\noindent\rule{\linewidth}{0.8pt}\par
% Original bottom caption retained; the active caption is above the table.
% \captionof{table}{The prompt template for target extraction quality assessment.}
% \label{tab:target_extraction_quality_assessment}
}

%% file: tables/tool_design_prompt.tex
{\scriptsize
\setlength{\parindent}{0pt}%
\setlength{\parskip}{0pt}%
\captionof{table}{The prompt template for tool design.}
\label{tab:tool_design}
\par\nopagebreak[4]\vspace{\baselineskip}
% \par\noindent\rule{\linewidth}{0.8pt}\par\vspace{2pt} % Original opening rule retained.
\par\noindent\rule{\linewidth}{0.8pt}\par\vspace{2pt}\nopagebreak[4]

You are an expert skilled at analyzing the quality of AI assistant responses.
I will provide you with a user instruction, a given constraint contained in that instruction, a response, and a list of counting objects for this constraint.
My goal is to determine whether the response satisfies the given constraint.
For this constraint, generate a Python auxiliary verification tool to help me make a more accurate judgment.
You must follow the principles below:

\vspace{\baselineskip}
\makebox[1.5em][l]{1.} Auxiliary verification means using a tool to automatically obtain important information from the response to assist a subsequent judge model to verify this constraint.
It does not mean using code to directly produce a judgment about whether the constraint is satisfied.

\makebox[1.5em][l]{2.}The generated auxiliary verification tool must satisfy the following requirements:

\hspace*{1.5em}\makebox[1.5em][l]{a.} The auxiliary verification tool should be generated only for completely objective constraints.
Subjective constraints must not be verified with code.
However, not every objective constraint necessarily requires an auxiliary verification tool.

\hspace*{1.5em}\makebox[1.5em][l]{b.}The information computed by the auxiliary verification tool must be fully discriminative and deterministic.
The computed information must differ significantly between responses that satisfy the constraint and responses that do not.

\hspace*{1.5em}\makebox[1.5em][l]{c.}The information computed by the auxiliary verification tool should cover the list of counting objects as comprehensively as possible without being limited to it.
An object in the list may be omitted if it is subjective or difficult to compute accurately with code.
You may also freely compute other information outside the list that you consider necessary.

\makebox[1.5em][l]{3.}The auxiliary verification tool must be a Python function satisfying the following requirements:

\hspace*{1.5em}\makebox[1.5em][l]{a.}You can freely design the function's input parameters, but their values must be response segments relevant to the constraint.

\hspace*{1.5em}\makebox[1.5em][l]{b.}The function must return a list.
Each item in the list must be a Dict containing one category of important information obtained by the auxiliary verification tool.
Each Dict must follow this format:

\texttt{\textasciigrave\textasciigrave\textasciigrave}

\texttt{\{}

\hspace*{1.5em}\texttt{"Description": ... (string; provide a detailed definition of the information obtained by the tool so that the judge model can understand and use it),}

\hspace*{1.5em}\texttt{"Computed Result": ... (any format; provide the computed information),}

\texttt{\}}

\texttt{\textasciigrave\textasciigrave\textasciigrave}

\hspace*{1.5em}\makebox[1.5em][l]{c.}All Python packages that need to be imported must be imported inside the function.
The function must not use any Python package that requires an additional download.

\makebox[1.5em][l]{4.}To ensure a consistent standard when counting words or characters or when splitting sentences, directly use the following functions implemented by human experts unless the constraint explicitly specifies otherwise:

\hspace*{1.5em}\makebox[1.5em][l]{a.}\texttt{count\_word(s: str) -> int}: Counts the words or Chinese characters in a string.
Its input is the string to be counted, and it returns the result as an integer.
Use the \texttt{len(s: str) -> int} function only when the constraint explicitly specifies a character count.

\hspace*{1.5em}\makebox[1.5em][l]{b.}\texttt{split\_sentences(s: str) -> List}: Splits a string into sentences.
Its input is the string, and it returns a list containing all sentences in the string.

\makebox[1.5em][l]{5.}If you choose to generate an auxiliary verification tool, you must also extract the corresponding value of each input parameter from the response according to the following principles:

\hspace*{1.5em}\makebox[1.5em][l]{a.}You should extract only continuous verbatim segments from the response.
You are strictly prohibited from modifying, adding, or deleting any content.

\hspace*{1.5em}\makebox[1.5em][l]{b.}If the extracted value corresponding to a parameter is the complete response, output \texttt{"ALL"} instead of reproducing the entire response.
If no corresponding value exists in the response, output \texttt{""}.

\vspace{\baselineskip}
Use the following output format:

\vspace{\baselineskip}
\texttt{\textasciigrave\textasciigrave\textasciigrave\textasciigrave}

\textbf{[The Start of Auxiliary Verification Tool]}

\texttt{\textasciigrave\textasciigrave\textasciigrave python}

... (Provide a Python tool that can assist in verifying the constraint according to the specifications above. If you choose not to generate an auxiliary verification tool, output ``None'' here.)

\texttt{\textasciigrave\textasciigrave\textasciigrave}

\textbf{[The End of Auxiliary Verification Tool]}

\vspace{\baselineskip}
\textbf{[The Start of Input Parameters]}

... (Provide the name, format, description, and value of every input parameter designed for the auxiliary verification tool in detail.
List the parameters in the same order as they appear in the function and use the JSON format below.
If you choose not to generate an auxiliary verification tool, output ``None'' here.)

\texttt{\textasciigrave\textasciigrave\textasciigrave json}

\texttt{[}

\hspace*{1.5em}\texttt{\{}

\hspace*{3em}\texttt{"Parameter Name": ... (the parameter name, which must match the name used in the auxiliary verification tool),}

\hspace*{3em}\texttt{"Format": ... (the parameter's data type),}

\hspace*{3em}\texttt{"Parameter Description": ... (the description and meaning of the parameter),}

\hspace*{3em}\texttt{"Parameter Value": ... (the value corresponding to the parameter in the response. It must be a complete verbatim segment from the response and must not be modified in any way. Its format must be consistent with the input parameter definition)}

\hspace*{1.5em}\texttt{\},}

\hspace*{1.5em}\texttt{...}

\texttt{]}

\texttt{\textasciigrave\textasciigrave\textasciigrave}

\textbf{[The End of Input Parameters]}

\texttt{\textasciigrave\textasciigrave\textasciigrave\textasciigrave}

\vspace{\baselineskip}
The user instruction, the corresponding constraint, the response, and the counting object list are provided below:

\textbf{[The Start of User Instruction]}

\textcolor{red}{\{instruction\}}

\textbf{[The End of User Instruction]}

\vspace{\baselineskip}
\textbf{[The Start of the Constraint]}

\textcolor{red}{\{constraint\}}

\textbf{[The End of the Constraint]}

\vspace{\baselineskip}
\textbf{[The Start of the Response]}

\textcolor{red}{\{response\}}

\textbf{[The End of the Response]}

\vspace{\baselineskip}
\textbf{[The Start of Counting Object List]}

\textcolor{red}{\{target\}}

\textbf{[The End of Counting Object List]}

% \par\vspace{2pt}\noindent\rule{\linewidth}{0.8pt}\par % Original closing rule retained.
\par\nopagebreak[4]\vspace{2pt}\noindent\rule{\linewidth}{0.8pt}\par
% Original bottom caption retained; the active caption is above the table.
% \captionof{table}{The prompt template for tool design.}
% \label{tab:tool_design}
}

%% file: tables/judge_prompt.tex
{\scriptsize
\setlength{\parindent}{0pt}%
\setlength{\parskip}{0pt}%
\captionof{table}{The prompt template for tool-grounded judgment.}
\label{tab:tool_grounded_judgement}
\par\nopagebreak[4]\vspace{\baselineskip}
% \par\noindent\rule{\linewidth}{0.8pt}\par\vspace{2pt} % Original opening rule retained.
\par\noindent\rule{\linewidth}{0.8pt}\par\vspace{2pt}\nopagebreak[4]

You are an expert skilled at evaluating the quality of responses generated by artificial intelligence assistants.
I will provide you with a user instruction, a constraint contained in that instruction, a response to the instruction, and a list of counting objects for this constraint.
You must carefully examine whether the response satisfies the corresponding constraint in the user instruction and explain your reason.
I will also provide some reference information (may be empty), which is automatically calculated by auxiliary verification tools for the corresponding constraint.
You should use this information cautiously when making your final judgment and provide a counting result for every counting object.
Note that auxiliary verification tools have limitations, so the reference information it provides is not guaranteed to be completely correct and may contain errors.
If you are confident that the reference information contains an obvious error, you may disregard it.
You must follow the principles below:

\vspace{\baselineskip}
1. Your judgment should be as strict as possible.
You should conclude ``[[The response satisfies the constraint]]'' only when the response completely satisfies every part of the corresponding constraint.
If the response contains any omission or error in fulfilling the constraint, you must conclude ``[[The response does not satisfy the constraint]]''.

2. Your judgment of the corresponding constraint must remain independent.
When evaluating the current constraint, you do not need to consider whether other parts of the user instruction are satisfied.

3. After providing your judgment, combine the reference information with your analysis and provide a counting result for every item in the counting object list.
Follow the principles below when providing the counting results:

\hspace*{1.5em} a. Pay attention to whether each counting object is ``Universal'' or ``Specific''.
For a specific counting object, use the original wording of that counting object.
For a universal counting object, do not directly use its original wording.
Instead, enumerate a separate counting result for every corresponding item that appears in the response.

\hspace*{1.5em} b. Provide counting results only for the items included in the counting object list.

4. For some constraints, the permissible bounds are not specified using a concrete number but are instead expressed through a relative quantitative relationship with a reference value, such as an inequality or arithmetic relationship.
In such cases, you must first calculate the reference value and then substitute it into the permissible bounds.

5. You must ensure that your conclusion and counting results are logically consistent.
If your judgment is ``[[The response satisfies the constraint]],'' the actual value of every counting result must fall within its permissible bounds range.
If your judgment is ``[[The response does not satisfy the constraint]],'' the actual value of at least one counting result must fall outside its permissible bounds range.

\vspace{\baselineskip}
You must strictly follow the format below when providing your analysis and judgment of the constraint:

\vspace{\baselineskip}
\texttt{\textasciigrave\textasciigrave\textasciigrave\textasciigrave}

\textbf{Constraint:} ... (Provide the original corresponding constraint verbatim, without making any modification)

\textbf{Analysis:} ... (Provide a detailed and meticulous analysis based on the specific content of the response, explaining whether the response satisfies the constraint)

\textbf{Conclusion:} ... (This must be either ``[[The response satisfies the constraint]]'' or ``[[The response does not satisfy the constraint]]'')

\textbf{Counting Results:}
\texttt{\textasciigrave\textasciigrave\textasciigrave json}

\texttt{[}

\hspace*{1.5em}\texttt{\{}

\hspace*{3em}\texttt{"Counting Object": ... (A string that describes the counting object in detail),}

\hspace*{3em}\texttt{"Permissible Bounds": ... (A two-element list specifying a closed interval in the form [a, b]. The two endpoints may be equal when only one value is valid. For qualitative constraints, the permissible bounds may be set to [0, 0] or [1, 1]),}

\hspace*{3em}\texttt{"Actual Value": ... (An integer specifying the actual value of this counting object in the response),}

\hspace*{1.5em}\texttt{\},}

\hspace*{1.5em}\texttt{...}

\texttt{]}

\texttt{\textasciigrave\textasciigrave\textasciigrave}

\texttt{\textasciigrave\textasciigrave\textasciigrave\textasciigrave}

\vspace{\baselineskip}
The user instruction, the corresponding constraint, the auxiliary verification results, the response, and the counting object list are provided below:

\textbf{[The Start of User Instruction]}

\textcolor{red}{\{instruction\}}

\textbf{[The End of User Instruction]}

\vspace{\baselineskip}
\textbf{[The Start of the Constraint]}

\textcolor{red}{\{constraint\}}

\textbf{[The End of the Constraint]}

\vspace{\baselineskip}
\textbf{[The Start of Auxiliary Verification Results]}

\textcolor{red}{\{verification\_results\}}

\textbf{[The End of Auxiliary Verification Results]}

\vspace{\baselineskip}
\textbf{[The Start of the Response]}

\textcolor{red}{\{response\}}

\textbf{[The End of the Response]}

\vspace{\baselineskip}
\textbf{[The Start of Counting Object List]}

\textcolor{red}{\{target\}}

\textbf{[The End of Counting Object List]}

% \par\vspace{2pt}\noindent\rule{\linewidth}{0.8pt}\par % Original closing rule retained.
\par\nopagebreak[4]\vspace{2pt}\noindent\rule{\linewidth}{0.8pt}\par
% Original bottom caption retained; the active caption is above the table.
% \captionof{table}{The prompt template for tool-grounded judgment.}
% \label{tab:tool_grounded_judgement}
}

%% file: custom.bib
@article{liu2023trustworthy,
  title={Trustworthy llms: a survey and guideline for evaluating large language models' alignment},
  author={Liu, Yang and Yao, Yuanshun and Ton, Jean-Francois and Zhang, Xiaoying and Guo, Ruocheng and Cheng, Hao and Klochkov, Yegor and Taufiq, Muhammad Faaiz and Li, Hang},
  journal={arXiv preprint arXiv:2308.05374},
  year={2023}
}

@article{lou2024large,
  title={Large language model instruction following: A survey of progresses and challenges},
  author={Lou, Renze and Zhang, Kai and Yin, Wenpeng},
  journal={Computational Linguistics},
  volume={50},
  number={3},
  pages={1053--1095},
  year={2024},
  publisher={MIT Press 255 Main Street, 9th Floor, Cambridge, Massachusetts 02142, USA~…}
}

@article{sun2024conifer,
  title={Conifer: Improving complex constrained instruction-following ability of large language models},
  author={Sun, Haoran and Liu, Lixin and Li, Junjie and Wang, Fengyu and Dong, Baohua and Lin, Ran and Huang, Ruohui},
  journal={arXiv preprint arXiv:2404.02823},
  year={2024}
}

@inproceedings{zhang-etal-2025-iopo,
    title = "{IOPO}: Empowering {LLM}s with Complex Instruction Following via Input-Output Preference Optimization",
    author = "Zhang, Xinghua  and
      Yu, Haiyang  and
      Fu, Cheng  and
      Huang, Fei  and
      Li, Yongbin",
    editor = "Che, Wanxiang  and
      Nabende, Joyce  and
      Shutova, Ekaterina  and
      Pilehvar, Mohammad Taher",
    booktitle = "Proceedings of the 63rd Annual Meeting of the Association for Computational Linguistics (Volume 1: Long Papers)",
    month = jul,
    year = "2025",
    address = "Vienna, Austria",
    publisher = "Association for Computational Linguistics",
    url = "https://aclanthology.org/2025.acl-long.1079/",
    doi = "10.18653/v1/2025.acl-long.1079",
    pages = "22185--22200",
    ISBN = "979-8-89176-251-0"
}

@inproceedings{
cheng2025spar,
title={{SP}aR: Self-Play with Tree-Search Refinement to Improve Instruction-Following in Large Language Models},
author={Jiale Cheng and Xiao Liu and Cunxiang Wang and Xiaotao Gu and Yida Lu and Dan Zhang and Yuxiao Dong and Jie Tang and Hongning Wang and Minlie Huang},
booktitle={The Thirteenth International Conference on Learning Representations},
year={2025},
url={https://openreview.net/forum?id=9chRqsPOGL}
}

@article{liu2025recast,
  title={RECAST: Strengthening LLMs' Complex Instruction Following with Constraint-Verifiable Data},
  author={Liu, Wenhao and Guo, Zhengkang and Xie, Mingchen and Xu, Jingwen and Huang, Zisu and Tian, Muzhao and Xu, Jianhan and Wu, Muling and Wang, Xiaohua and Lv, Changze and others},
  journal={arXiv preprint arXiv:2505.19030},
  year={2025}
}

@inproceedings{peng-etal-2025-verif,
    title = "{V}er{IF}: Verification Engineering for Reinforcement Learning in Instruction Following",
    author = "Peng, Hao  and
      Qi, Yunjia  and
      Wang, Xiaozhi  and
      Xu, Bin  and
      Hou, Lei  and
      Li, Juanzi",
    editor = "Christodoulopoulos, Christos  and
      Chakraborty, Tanmoy  and
      Rose, Carolyn  and
      Peng, Violet",
    booktitle = "Proceedings of the 2025 Conference on Empirical Methods in Natural Language Processing",
    month = nov,
    year = "2025",
    address = "Suzhou, China",
    publisher = "Association for Computational Linguistics",
    url = "https://aclanthology.org/2025.emnlp-main.1542/",
    doi = "10.18653/v1/2025.emnlp-main.1542",
    pages = "30324--30339",
    ISBN = "979-8-89176-332-6"
}

@article{zhao2023survey,
  title={A survey of large language models},
  author={Zhao, Wayne Xin and Zhou, Kun and Li, Junyi and Tang, Tianyi and Wang, Xiaolei and Hou, Yupeng and Min, Yingqian and Zhang, Beichen and Zhang, Junjie and Dong, Zican and others},
  journal={arXiv preprint arXiv:2303.18223},
  volume={1},
  number={2},
  year={2023}
}

@article{pyatkin2026generalizing,
  title={Generalizing verifiable instruction following},
  author={Pyatkin, Valentina and Malik, Saumya and Graf, Victoria and Ivison, Hamish and Huang, Shengyi and Dasigi, Pradeep and Lambert, Nathan and Hajishirzi, Hanna},
  journal={Advances in Neural Information Processing Systems},
  volume={38},
  year={2026}
}

@inproceedings{ye2026muldimif,
  title={MulDimIF: A Multi-Dimensional Constraint Framework for Evaluating and Improving Instruction Following in Large Language Models},
  author={Ye, Junjie and Huang, Caishuang and Chen, Zhuohan and Fu, Wenjie and Yang, Chenyuan and Yang, Leyi and Wu, Yilong and Wang, Peng and Zhou, Meng and Yang, Xiaolong and others},
  booktitle={Findings of the Association for Computational Linguistics: ACL 2026},
  pages={2078--2104},
  year={2026}
}

@article{huang2024survey,
  title={A survey of safety and trustworthiness of large language models through the lens of verification and validation},
  author={Huang, Xiaowei and Ruan, Wenjie and Huang, Wei and Jin, Gaojie and Dong, Yi and Wu, Changshun and Bensalem, Saddek and Mu, Ronghui and Qi, Yi and Zhao, Xingyu and others},
  journal={Artificial Intelligence Review},
  volume={57},
  number={7},
  pages={175},
  year={2024},
  publisher={Springer}
}

@article{mao2026ifhierbench,
  title={IFHierBench: Hierarchical Instruction Following for Large Language Models},
  author={Mao, Yuetian and Chen, Chunyang},
  journal={arXiv preprint arXiv:2607.27912},
  year={2026}
}

@article{yang2025qwen3,
  title={Qwen3 technical report},
  author={Yang, An and Li, Anfeng and Yang, Baosong and Zhang, Beichen and Hui, Binyuan and Zheng, Bo and Yu, Bowen and Gao, Chang and Huang, Chengen and Lv, Chenxu and others},
  journal={arXiv preprint arXiv:2505.09388},
  year={2025}
}

@article{guo2025deepseek,
  title={DeepSeek-R1 incentivizes reasoning in LLMs through reinforcement learning},
  author={Guo, Daya and Yang, Dejian and Zhang, Haowei and Song, Junxiao and Wang, Peiyi and Zhu, Qihao and Xu, Runxin and Zhang, Ruoyu and Ma, Shirong and Bi, Xiao and others},
  journal={Nature},
  volume={645},
  number={8081},
  pages={633--638},
  year={2025},
  publisher={Nature Publishing Group UK London}
}

@article{qin2026incentivizing,
  title={Incentivizing reasoning for advanced instruction-following of large language models},
  author={Qin, Yulei and Li, Gang and Li, Zongyi and Xu, Zihan and Shi, Yuchen and Lin, Zhekai and Cui, Xiao and Li, Ke and Sun, Xing},
  journal={Advances in Neural Information Processing Systems},
  volume={38},
  pages={108337--108401},
  year={2026}
}

@article{zhang2025replay,
  title={Replay failures as successes: Sample-efficient reinforcement learning for instruction following},
  author={Zhang, Kongcheng and Yao, Qi and Liu, Shunyu and Zhang, Wenjian and Cen, Min and Zhou, Yang and Fang, Wenkai and Zhao, Yiru and Lai, Baisheng and Song, Mingli},
  journal={arXiv preprint arXiv:2512.23457},
  year={2025}
}

@inproceedings{he-etal-2024-complex,
    title = "From Complex to Simple: Enhancing Multi-Constraint Complex Instruction Following Ability of Large Language Models",
    author = "He, Qianyu  and
      Zeng, Jie  and
      He, Qianxi  and
      Liang, Jiaqing  and
      Xiao, Yanghua",
    editor = "Al-Onaizan, Yaser  and
      Bansal, Mohit  and
      Chen, Yun-Nung",
    booktitle = "Findings of the Association for Computational Linguistics: EMNLP 2024",
    month = nov,
    year = "2024",
    address = "Miami, Florida, USA",
    publisher = "Association for Computational Linguistics",
    url = "https://aclanthology.org/2024.findings-emnlp.637/",
    doi = "10.18653/v1/2024.findings-emnlp.637",
    pages = "10864--10882"
}

@article{zhang2026lifebench,
  title={Lifebench: Evaluating length instruction following in large language models},
  author={Zhang, Wei and Zhou, Zhenhong and Wang, Kun and Fang, Junfeng and Xu, Rongwu and Zhang, Yuanhe and Wang, Rui and Zhang, Ge and Li, Xinfeng and Sun, Li and others},
  journal={Advances in Neural Information Processing Systems},
  volume={38},
  year={2026}
}

@inproceedings{xia2024fofo,
  title={FOFO: A benchmark to evaluate LLMs’ format-following capability},
  author={Xia, Congying and Xing, Chen and Du, Jiangshu and Yang, Xinyi and Feng, Yihao and Xu, Ran and Yin, Wenpeng and Xiong, Caiming},
  booktitle={Proceedings of the 62nd Annual Meeting of the Association for Computational Linguistics (Volume 1: Long Papers)},
  pages={680--699},
  year={2024}
}

@article{zhang2026larft,
  title={LARFT: Closing the Cognition-Action Gap for Length Instruction Following in Large Language Models},
  author={Zhang, Wei and Du, Lintong and Zhang, Yuanhe and Zhou, Zhenhong and Wang, Kun and Sun, Li and Su, Sen},
  journal={arXiv preprint arXiv:2603.19255},
  year={2026}
}

@inproceedings{jie2024prompt,
  title={Prompt-based length controlled generation with multiple control types},
  author={Jie, Renlong and Meng, Xiaojun and Shang, Lifeng and Jiang, Xin and Liu, Qun},
  booktitle={Findings of the Association for Computational Linguistics: ACL 2024},
  pages={1067--1085},
  year={2024}
}

@inproceedings{yao2025reff,
  title={Reff: Reinforcing format faithfulness in language models across varied tasks},
  author={Yao, Jiashu and Huang, Heyan and Liu, Zeming and Wen, Haoyu and Su, Wei and Qian, Boao and Guo, Yuhang},
  booktitle={Proceedings of the AAAI Conference on Artificial Intelligence},
  volume={39},
  number={24},
  pages={25660--25668},
  year={2025}
}

@inproceedings{wang2025verifiable,
  title={Verifiable format control for large language model generations},
  author={Wang, Zhaoyang and Jiang, Jinqi and Zhou, Huichi and Zheng, Wenhao and Zhang, Xuchao and Bansal, Chetan and Yao, Huaxiu},
  booktitle={Findings of the Association for Computational Linguistics: NAACL 2025},
  pages={3499--3513},
  year={2025}
}

@article{zhou2023instruction,
  title={Instruction-following evaluation for large language models},
  author={Zhou, Jeffrey and Lu, Tianjian and Mishra, Swaroop and Brahma, Siddhartha and Basu, Sujoy and Luan, Yi and Zhou, Denny and Hou, Le},
  journal={arXiv preprint arXiv:2311.07911},
  year={2023}
}

@inproceedings{qi2025constraint,
  title={Constraint back-translation improves complex instruction following of large language models},
  author={Qi, Yunjia and Peng, Hao and Wang, Xiaozhi and Xu, Bin and Hou, Lei and Li, Juanzi},
  booktitle={Proceedings of the 34th ACM International Conference on Information and Knowledge Management},
  pages={2388--2398},
  year={2025}
}

@article{shao2024deepseekmath,
  title={Deepseekmath: Pushing the limits of mathematical reasoning in open language models},
  author={Shao, Zhihong and Wang, Peiyi and Zhu, Qihao and Xu, Runxin and Song, Junxiao and Bi, Xiao and Zhang, Haowei and Zhang, Mingchuan and Li, YK and Wu, Yang and others},
  journal={arXiv preprint arXiv:2402.03300},
  year={2024}
}

@inproceedings{lightman2024let,
  title={Let's verify step by step},
  author={Lightman, Hunter and Kosaraju, Vineet and Burda, Yuri and Edwards, Harrison and Baker, Bowen and Lee, Teddy and Leike, Jan and Schulman, John and Sutskever, Ilya and Cobbe, Karl},
  booktitle={International Conference on Learning Representations},
  volume={2024},
  pages={39578--39601},
  year={2024}
}

@article{rein2023gpqa,
  title={Gpqa: A graduate-level google-proof q\&a benchmark},
  author={Rein, David and Hou, Betty Li and Stickland, Asa Cooper and Petty, Jackson and Pang, Richard Yuanzhe and Dirani, Julien and Michael, Julian and Bowman, Samuel R},
  journal={arXiv preprint arXiv:2311.12022},
  year={2023}
}

@article{wang2024mmlu,
  title={Mmlu-pro: A more robust and challenging multi-task language understanding benchmark},
  author={Wang, Yubo and Ma, Xueguang and Zhang, Ge and Ni, Yuansheng and Chandra, Abhranil and Guo, Shiguang and Ren, Weiming and Arulraj, Aaran and He, Xuan and Jiang, Ziyan and others},
  journal={Advances in Neural Information Processing Systems},
  volume={37},
  pages={95266--95290},
  year={2024}
}

@inproceedings{zheng2024llamafactory,
  title={Llamafactory: Unified efficient fine-tuning of 100+ language models},
  author={Zheng, Yaowei and Zhang, Richong and Zhang, Junhao and Ye, Yanhan and Luo, Zheyan},
  booktitle={Proceedings of the 62nd annual meeting of the association for computational linguistics (volume 3: system demonstrations)},
  pages={400--410},
  year={2024}
}

@inproceedings{sheng2025hybridflow,
  title={Hybridflow: A flexible and efficient rlhf framework},
  author={Sheng, Guangming and Zhang, Chi and Ye, Zilingfeng and Wu, Xibin and Zhang, Wang and Zhang, Ru and Peng, Yanghua and Lin, Haibin and Wu, Chuan},
  booktitle={Proceedings of the Twentieth European Conference on Computer Systems},
  pages={1279--1297},
  year={2025}
}

@article{rafailov2023direct,
  title={Direct preference optimization: Your language model is secretly a reward model},
  author={Rafailov, Rafael and Sharma, Archit and Mitchell, Eric and Manning, Christopher D and Ermon, Stefano and Finn, Chelsea},
  journal={Advances in neural information processing systems},
  volume={36},
  pages={53728--53741},
  year={2023}
}

@misc{seed2026seed,
  title={Seed 2.0 Official Launch},
  author={{Seed Team}},
  year={2026},
  url={https://seed.bytedance.com/en/blog/seed-2-0-official-launch},
}

@inproceedings{rajbhandari2020zero,
  title={Zero: Memory optimizations toward training trillion parameter models},
  author={Rajbhandari, Samyam and Rasley, Jeff and Ruwase, Olatunji and He, Yuxiong},
  booktitle={SC20: International Conference for High Performance Computing, Networking, Storage and Analysis},
  pages={1--16},
  year={2020},
  organization={IEEE}
}

@inproceedings{rasley2020deepspeed,
  title={Deepspeed: System optimizations enable training deep learning models with over 100 billion parameters},
  author={Rasley, Jeff and Rajbhandari, Samyam and Ruwase, Olatunji and He, Yuxiong},
  booktitle={Proceedings of the 26th ACM SIGKDD international conference on knowledge discovery \& data mining},
  pages={3505--3506},
  year={2020}
}

@article{kingma2014adam,
  title={Adam: A method for stochastic optimization},
  author={Kingma, Diederik P and Ba, Jimmy},
  journal={arXiv preprint arXiv:1412.6980},
  year={2014}
}

@article{hou2024chatglm,
  title={Chatglm-rlhf: Practices of aligning large language models with human feedback},
  author={Hou, Zhenyu and Niu, Yilin and Du, Zhengxiao and Zhang, Xiaohan and Liu, Xiao and Zeng, Aohan and Zheng, Qinkai and Huang, Minlie and Wang, Hongning and Tang, Jie and others},
  journal={arXiv preprint arXiv:2404.00934},
  year={2024}
}

@inproceedings{kwon2023efficient,
  title={Efficient memory management for large language model serving with pagedattention},
  author={Kwon, Woosuk and Li, Zhuohan and Zhuang, Siyuan and Sheng, Ying and Zheng, Lianmin and Yu, Cody Hao and Gonzalez, Joseph and Zhang, Hao and Stoica, Ion},
  booktitle={Proceedings of the 29th Symposium on Operating Systems Principles},
  pages={611--626},
  year={2023}
}

@article{lambert2024t,
  title={T$\backslash$" ulu 3: Pushing frontiers in open language model post-training},
  author={Lambert, Nathan and Morrison, Jacob and Pyatkin, Valentina and Huang, Shengyi and Ivison, Hamish and Brahman, Faeze and Miranda, Lester James V and Liu, Alisa and Dziri, Nouha and Lyu, Shane and others},
  journal={arXiv preprint arXiv:2411.15124},
  year={2024}
}

@inproceedings{yang-etal-2026-imprif,
    title = "{I}mp{RIF}: Stronger Implicit Reasoning Leads to Better Complex Instruction Following",
    author = "Yang, Yuancheng  and
      Yang, Lin  and
      Wang, Xu  and
      Tong, Chao  and
      Yang, Haihua",
    editor = "Liakata, Maria  and
      Moreira, Viviane P.  and
      Zhang, Jiajun  and
      Jurgens, David",
    booktitle = "Proceedings of the 64th Annual Meeting of the {A}ssociation for {C}omputational {L}inguistics (Volume 1: Long Papers)",
    month = jul,
    year = "2026",
    address = "San Diego, California, United States",
    publisher = "Association for Computational Linguistics",
    url = "https://aclanthology.org/2026.acl-long.1796/",
    doi = "10.18653/v1/2026.acl-long.1796",
    pages = "38771--38796",
    ISBN = "979-8-89176-390-6"
}

@article{wen2024benchmarking,
  title={Benchmarking complex instruction-following with multiple constraints composition},
  author={Wen, Bosi and Ke, Pei and Gu, Xiaotao and Wu, Lindong and Huang, Hao and Zhou, Jinfeng and Li, Wenchuang and Hu, Binxin and Gao, Wendy and Xu, Jiaxin and others},
  journal={Advances in Neural Information Processing Systems},
  volume={37},
  pages={137610--137645},
  year={2024}
}

@inproceedings{wen-etal-2026-rewardbench,
    title = "{IF}-{R}eward{B}ench: Benchmarking Judge Models for Instruction-Following Evaluation",
    author = "Wen, Bosi  and
      Niu, Yilin  and
      Wang, Cunxiang  and
      Ling, Xiaoying  and
      Zhang, Ying  and
      Ke, Pei  and
      Wang, Hongning  and
      Huang, Minlie",
    editor = "Liakata, Maria  and
      Moreira, Viviane P.  and
      Zhang, Jiajun  and
      Jurgens, David",
    booktitle = "Proceedings of the 64th Annual Meeting of the {A}ssociation for {C}omputational {L}inguistics (Volume 1: Long Papers)",
    month = jul,
    year = "2026",
    address = "San Diego, California, United States",
    publisher = "Association for Computational Linguistics",
    url = "https://aclanthology.org/2026.acl-long.1092/",
    doi = "10.18653/v1/2026.acl-long.1092",
    pages = "23816--23843",
    ISBN = "979-8-89176-390-6"
}

@inproceedings{zhu2018texygen,
  title={Texygen: A benchmarking platform for text generation models},
  author={Zhu, Yaoming and Lu, Sidi and Zheng, Lei and Guo, Jiaxian and Zhang, Weinan and Wang, Jun and Yu, Yong},
  booktitle={The 41st international ACM SIGIR conference on research \& development in information retrieval},
  pages={1097--1100},
  year={2018}
}

@inproceedings{li2016diversity,
  title={A diversity-promoting objective function for neural conversation models},
  author={Li, Jiwei and Galley, Michel and Brockett, Chris and Gao, Jianfeng and Dolan, William B},
  booktitle={Proceedings of the 2016 conference of the North American chapter of the association for computational linguistics: human language technologies},
  pages={110--119},
  year={2016}
}

@article{zhang2018generating,
  title={Generating informative and diverse conversational responses via adversarial information maximization},
  author={Zhang, Yizhe and Galley, Michel and Gao, Jianfeng and Gan, Zhe and Li, Xiujun and Brockett, Chris and Dolan, Bill},
  journal={Advances in Neural Information Processing Systems},
  volume={31},
  year={2018}
}

@inproceedings{peng-etal-2025-agentic,
    title = "Agentic Reward Modeling: Integrating Human Preferences with Verifiable Correctness Signals for Reliable Reward Systems",
    author = "Peng, Hao  and
      Qi, Yunjia  and
      Wang, Xiaozhi  and
      Yao, Zijun  and
      Xu, Bin  and
      Hou, Lei  and
      Li, Juanzi",
    editor = "Che, Wanxiang  and
      Nabende, Joyce  and
      Shutova, Ekaterina  and
      Pilehvar, Mohammad Taher",
    booktitle = "Proceedings of the 63rd Annual Meeting of the Association for Computational Linguistics (Volume 1: Long Papers)",
    month = jul,
    year = "2025",
    address = "Vienna, Austria",
    publisher = "Association for Computational Linguistics",
    url = "https://aclanthology.org/2025.acl-long.775/",
    doi = "10.18653/v1/2025.acl-long.775",
    pages = "15934--15949",
    ISBN = "979-8-89176-251-0"
}

@article{loshchilov2017decoupled,
  title={Decoupled weight decay regularization},
  author={Loshchilov, Ilya and Hutter, Frank},
  journal={arXiv preprint arXiv:1711.05101},
  year={2017}
}

@inproceedings{wen2026if,
    title = "{IF}-{CRITIC}: Towards a Fine-Grained {LLM} Critic for Instruction-Following Evaluation",
    author = "Wen, Bosi  and
      Niu, Yilin  and
      Wang, Cunxiang  and
      Ke, Pei  and
      Ling, Xiaoying  and
      Zhang, Ying  and
      Zeng, Aohan  and
      Wang, Hongning  and
      Huang, Minlie",
    editor = "Liakata, Maria  and
      Moreira, Viviane P.  and
      Zhang, Jiajun  and
      Jurgens, David",
    booktitle = "Proceedings of the 64th Annual Meeting of the {A}ssociation for {C}omputational {L}inguistics (Volume 1: Long Papers)",
    month = jul,
    year = "2026",
    address = "San Diego, California, United States",
    publisher = "Association for Computational Linguistics",
    url = "https://aclanthology.org/2026.acl-long.1147/",
    doi = "10.18653/v1/2026.acl-long.1147",
    pages = "25004--25029",
    ISBN = "979-8-89176-390-6"
}
